\documentclass[runningheads]{llncs}

\usepackage{eccv}
\def\RefAppendix{1}

\usepackage{eccvabbrv}

\usepackage{graphicx}
\usepackage{amsmath}
\usepackage{amssymb}
\usepackage{booktabs}
\usepackage{multirow}
\usepackage{makecell}
\usepackage{enumitem} %
\usepackage{soul}
\usepackage{changepage,threeparttable}
\usepackage[accsupp]{axessibility}
\usepackage{mmstyle}

\definecolor{MyRed}{HTML}{FF0001}

\newcommand{\tablestyle}[2]{\setlength{\tabcolsep}{#1}\renewcommand{\arraystretch}{#2}\centering\small}

\usepackage[misc]{ifsym}

\usepackage{graphicx}
\usepackage{booktabs}

\usepackage[accsupp]{axessibility}  %

\usepackage[capbesideposition=outside,capbesidesep=quad]{floatrow}

\DeclareFloatVCode{beforefloat}{\vspace{-2pt}}
\DeclareFloatVCode{afterfloat}{\vspace{-2pt}}
\usepackage[pagebackref,breaklinks,colorlinks,citecolor=eccvblue]{hyperref}

\usepackage{orcidlink}

\usepackage{pifont}
\usepackage{algpseudocode}
\usepackage{algorithm}

\usepackage[capitalize]{cleveref}
\crefname{section}{Sec.}{Secs.}
\Crefname{section}{Section}{Sections}
\Crefname{table}{Table}{Tables}
\crefname{table}{Tab.}{Tabs.}

\newcommand{\boldparagraph}[1]{\vspace{3pt}\noindent{\bf #1}}
\newcommand{\ours}{T-MAE\xspace}  %
\newcommand{\ourFusion}{WCA\xspace}  %
\newcommand{\ourBackbone}{SiamWCA\xspace}  %

\newcommand{\beginsupplement}{%
        \setcounter{table}{0}
        \renewcommand{\thetable}{S\arabic{table}}%
        \setcounter{figure}{0}
        \renewcommand{\thefigure}{S\arabic{figure}}%
        \setcounter{section}{0}
        \renewcommand{\thesection}{S\arabic{section}}%
     }

\newcommand{\red}[1]{{\color{red}#1}}

\definecolor{colorFst}{HTML}{bde6cd}       %
\definecolor{colorSnd}{HTML}{e4eebc}       %
\definecolor{colorTrd}{HTML}{fff8c5}       %

\newcommand{\fs}{\cellcolor{colorFst}\bf}   %
\newcommand{\nd}{\cellcolor{colorSnd}}      %
\newcommand{\rd}{\cellcolor{colorTrd}}      %

\definecolor{mgreen}{RGB}{0,176,80}
\newcommand\up[1]{\textcolor{red}{$^{\uparrow{#1}}$}}

\newcommand{\greencheck}{{\color{PineGreen}\checkmark}}
\newcommand{\redx}{{\color{red}\ding{55}}}

\begin{document}

\title{{\ours}: Temporal Masked Autoencoders for Point Cloud Representation Learning}

\titlerunning{T-MAE}

\author{Weijie Wei\orcidlink{0000-0002-5952-5341} \and
Fatemeh Karimi Nejadasl \and
Theo Gevers \and
Martin R. Oswald\orcidlink{0000-0002-1183-9958}}

\authorrunning{W.~Wei et al.}

\institute{University of Amsterdam, the Netherlands}

\maketitle

\begin{abstract}
The scarcity of annotated data in LiDAR point cloud understanding hinders effective representation learning. Consequently, scholars have been actively investigating efficacious self-supervised pre-training paradigms.
Nevertheless, temporal information, which is inherent in the LiDAR point cloud sequence, is consistently disregarded.
To better utilize this property, we propose an effective pre-training strategy, namely Temporal Masked Auto-Encoders (T-MAE), which takes as input temporally adjacent frames and learns temporal dependency.
A SiamWCA backbone, containing a Siamese encoder and a windowed cross-attention (WCA) module, is established for the two-frame input.
Considering that the movement of an ego-vehicle alters the view of the same instance, temporal modeling also serves as a robust and natural data augmentation, enhancing the comprehension of target objects.
\ourBackbone is a powerful architecture but heavily relies on annotated data.
Our \ours pre-training strategy alleviates its demand for annotated data.
Comprehensive experiments demonstrate that \ours achieves the best performance on both Waymo and ONCE datasets among competitive self-supervised approaches.
Codes will be released \hyperlink{https://github.com/codename1995/T-MAE}{here}.
\end{abstract}

\section{Introduction}
\label{sec:intro}

\begin{figure*}[t] 
    \centering
    \scriptsize
    \setlength{\tabcolsep}{0pt}
    \floatbox[\capbeside]{figure}[0.54\columnwidth]%
    {\caption{\textbf{\ours performance on Waymo~\cite{pSun_2020_Waymo}.}
    \textbf{Left:} Each point triplet shows the performance differences to three models finetuned with the same data.
    The triplets show finetuned models with 8K, 16K, 32K labeled frames (left to right).
    Our \ours pre-training outperforms both random initialization and the SOTA SSL method MV-JAR~\cite{xu_mv-jar_2023} with significantly fewer iterations.
    \textbf{Right:} \ours yields higher mAPH for pedestrians when finetuned with half the labeled data than MV-JAR.}
    \label{fig:fig1}
    }{
    \begin{tabular}{cc}
    \vspace{-5pt}
      \includegraphics[width=0.5\linewidth]{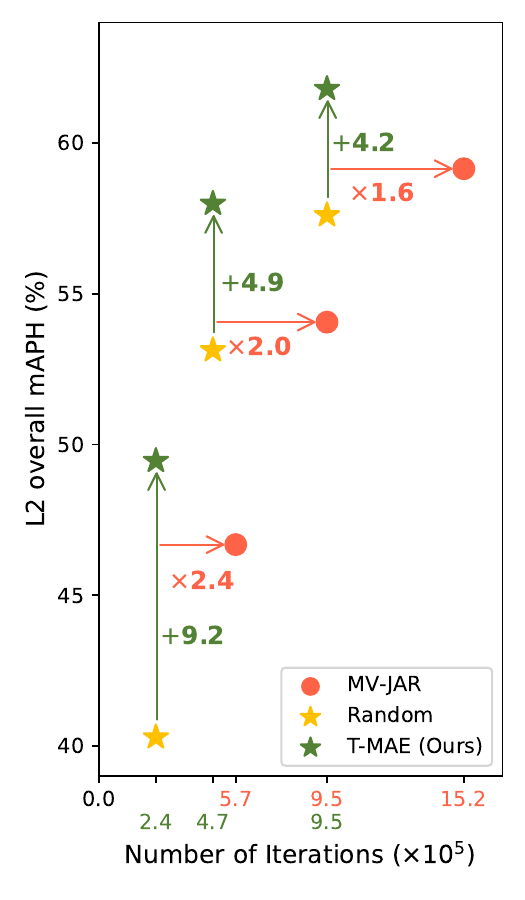} &
      \includegraphics[width=0.5\linewidth]{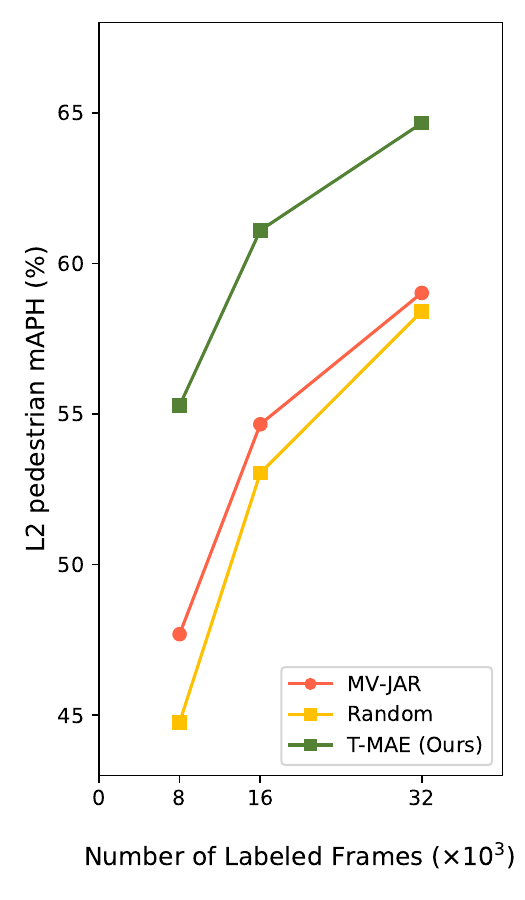}
    \end{tabular}}
\end{figure*}

As deep neural networks become more complex, the available amount of labeled data is often insufficient to adequately train huge models~\cite{jDevlin_2019_BERT, kHe_2021_MAE}, \eg, Vision Transformer (ViT)~\cite{aDosovitskiy_2021_ViT}.
Consequently, there is a growing interest in exploring self-supervised learning (SSL) approaches as a potential solution to overcome this limitation.
SSL serves as a pre-training technique with unlabeled data, accelerating the convergence of the models and improving their performance for downstream tasks~\cite{jbGrill_2020_BYOL, mCaron_2021_DINO, xChen_2021_MoCoV3}.
The same challenge extends to the domain of point clouds, where annotations are more costly and time-consuming to obtain~\cite{aXiao_2023_PCSSL_Survey, bFei_2023_PCSSL_Survey}.
For instance, a mere 10\% and 0.8\% of frames are annotated in the nuScene~\cite{nuscenes} and ONCE~\cite{jMao_2021_ONCE} datasets, respectively.
This challenge makes pre-training for point cloud understanding a non-trivial endeavour.

Prior works mainly focus on synthetic and isolated objects~\cite{xYan_2022_PointCMT, rZhang_2022_Point-M2AE, xYu_2022_Point-BERT, yPang_2022_PointMAE, hLiu_2022_MaskPoint, jJiang_2023_MAE3D, rZhang_2023_I2P-MAE} and indoor scene understanding~\cite{jHou_2021_ExploringDataEfficient3D, lLi_2022_CloserLookInvariances, rYamada_2022_PointCloudPretraining, yRao_2021_RandomRooms, lLiu_2023_CPCM}.
Transferring these methods to outdoor LiDAR points is challenging due to their sparsity and dynamic environmental conditions.
At present, most of the self-supervised methods used for understanding point clouds in autonomous driving rely on contrastive learning~\cite{sXie_2020_PointContrast, jYin_2022_ProposalContrast, nunes_segcontrast_2022, bPang_2023_Unsupervised3DPoint, nunes_tarl}.
These approaches model the similarity and dissimilarity between entities, such as segments~\cite{nunes_tarl, yWu_2023_SpatiotemporalSelfSupervisedLearning} and/or points~\cite{sXie_2020_PointContrast}.
In the wake of masked image modeling as a pretext task~\cite{kHe_2021_MAE}, efforts have also been devoted to the reconstruction of masked points~\cite{OccupancyMAE, tian_geomae_2023, yang_gd-mae_2023, xu_mv-jar_2023}.
The main idea is randomly masking points or voxels and urging the network to infer the coordinates of points~\cite{yang_gd-mae_2023} and/or voxels~\cite{xu_mv-jar_2023} or other properties, \eg, occupancy~\cite{OccupancyMAE, aBoulch_2023_ALSO} and curvature~\cite{tian_geomae_2023}.
Nevertheless, these methodologies often operate within the confines of a single-frame scenario, disregarding the fact that LiDAR data is typically acquired on a frame-by-frame basis.
In other words, the valuable semantic information in temporally adjacent frames is barely exploited.

Several methods attempt to leverage temporal information~\cite{sHuang_2021_STRL, yWu_2023_SpatiotemporalSelfSupervisedLearning, nunes_tarl, hLiang_2021_ExploringGeometryawareContrast} by incorporating multi-frame input during the self-supervised phase but their core concepts remain grounded in contrastive learning.
Specifically, the point clouds captured at different times are treated as augmented samples of the same scene, without including temporal correspondence into the modeling procedure.

Therefore, we propose a new self-supervised paradigm, namely \ours, to exploit the accumulated observations. 
During the pre-training stage, the current scan is voxelized with a high masking ratio, while the previous scan is fed entirely to the encoder. 
Then, the pretext task is to reconstruct the current scan by incorporating voxel embeddings of the past scan, visible voxel embeddings of the current scan, and the position of masked voxels. 
This way, the proposed windowed cross-attention module learns to incorporate historical information into the current frame using unlabeled data. The \ours pre-training strategy endows the network with both a powerful representation for sparse point clouds and the capacity to strengthen the present by learning from the past. As shown in Fig.~\ref{fig:fig1}, the proposed \ours achieves higher overall and pedestrian-specific mAPH than the randomly initialized baseline, namely the same model but trained from scratch. Moreover, \ours also outperforms state-of-the-art MV-JAR~\cite{xu_mv-jar_2023} with over $1.6 \times$ to $2.4 \times$ fewer finetuning iterations. 

Our \textbf{contributions} are summarized as follows: 
\textbf{1)} We propose \ours, a novel and effective SSL approach for representation learning of sparse point clouds, that learns temporal modeling in the process of reconstructing masked points. 
\textbf{2)} We design a SiamWCA backbone, containing a Siamese encoder and a windowed sparse cross-attention (WCA) module, to incorporate historical information. 
\textbf{3)} Our experiments demonstrate the efficacy of \ours by attaining substantial improvements on the Waymo and ONCE datasets.
Notably, \ours with 5\% labeled data outperforms the SOTA SSL approach MV-JAR in terms of mAPH for pedestrians, even if MV-JAR employs 10\% labeled data.

\section{Related Work}
\label{sec:related_works}

\boldparagraph{Static Self-supervised Learning.}
SSL for point clouds is a burgeoning field due to the scarcity of annotations. 
In general, the pipeline of SSL consists of two phases. 
First, the network is trained using a pretext task with unlabeled data.
Second, in downstream tasks such as segmentation and detection, the pre-trained weights are loaded to the backbone, and the backbone is attached with task-specific heads. 
The resulting model is finetuned with annotated data.

The initial research efforts are primarily directed towards contrastive learning.
The underlying hypothesis is that an image or 3D scene demonstrates feature equivalence even after undergoing different transformations~\cite{xChen_2021_MoCoV3, mCaron_2021_SwAV}.
One of the key challenges in the point cloud domain is establishing correspondences across scenes.
PointContrast~\cite{sXie_2020_PointContrast} and DepthContrast~\cite{zZhang_2021_DepthContrast} track the points while performing different transformations. GCC-3D~\cite{hLiang_2021_ExploringGeometryawareContrast} exploit sequential information to obtain pseudo instances and then perform contrasting on the instance level. 
SegContrast~\cite{nunes_segcontrast_2022} firstly obtains segments by an unsupervised clustering and then contrasts segments between two transformed views. 
These methods allow the network to learn equivalence regarding geometric transformations. 
However, contrastive SSL approaches usually suffer from careful tuning of hyperparameters and complicated pre- or post-processing to find correspondences.

The focus has shifted to reconstructing masked points, following the success of masked image modeling~\cite{kHe_2021_MAE, cWei_2022_MaskFeat}.
The masking strategies remain consistent, \ie voxelize a point cloud into cubes followed by random masking of these cubes. The reconstruction targets vary across papers.
OccupancyMAE~\cite{OccupancyMAE} classifies voxels to whether they are occupied. Geo-MAE~\cite{tian_geomae_2023} infers occupancy as well as the normal and curvature of each masked voxel. The GD-MAE's pretext task is to reconstruct a fixed number of points for masked voxels~\cite{yang_gd-mae_2023}. MV-JAR~\cite{xu_mv-jar_2023} partitions the masked Voxel into two categories, necessitating the network to either predict the coordinates of the voxels or produce the points themselves. In this paper, we employ the same masking strategy as GD-MAE~\cite{yang_gd-mae_2023} but expand the knowledge source to include both the current visible voxels and voxels from a previous frame as a reference.

\boldparagraph{Temporal Self-supervised Learning.}
There are many SSL methods designed to address temporal or spatiotemporal tasks in the video domain. MAE-ST~\cite{cFeichtenhofer_2022_MAE-ST} employs a random patch masking strategy over consecutive frames. The model is tasked with recovering these masked patches while considering information from adjacent frames. VideoMAE~\cite{zTong_2022_VideoMAE} retains the same reconstruction pipeline as the pretext task but exploits a tube masking strategy. SiameseMAE~\cite{aGupta_2023_SiameseMAE} learns object-centred representations with the help of cross-attention layers and an asymmetrical masking technique on consecutive frames. There are also content-based masking strategies, \eg, motion-guided masking~\cite{bHuang_2023_MGMAE, yMao_2023_MaskedMotionPredictors}. The primary concept underlying these methods is acquiring an understanding of temporal dependency across frames through the process of reconstructing patches with reference to consecutive frames. 

While this concept is effectively used in video understanding, it is rarely applied to learn sparse point cloud representations. This is because point clouds are typically handled on a frame-by-frame basis rather than being regarded as a temporal sequence. For instance, when using temporally adjacent scans as input during self-supervised learning, both STSSL~\cite{yWu_2023_SpatiotemporalSelfSupervisedLearning} and TARL~\cite{nunes_tarl} use HDBSCAN~\cite{rjgbCampello_2013_HDBSCAN} to obtain segments and apply self-supervision by minimizing the feature distance between the same segments or points of neighbour frames. While they achieve impressive results, two issues persist 1) The use of elaborate and time-consuming pre-processing methods to obtain segments. 2) A primary focus on object consistency across frames rather than understanding object motion.

Unlike other approaches, we aim to reduce the reliance on complex preprocessing techniques and focus on enabling the model to establish temporal correspondence through self-supervised learning from unlabeled data.

\section{Method} \label{sec:method}
We first briefly discuss important preliminaries in Sec.~\ref{sec:preliminaries} from previous works although they mainly focus on the single-frame setting.
To incorporate historical frames, we build up our framework as in Fig.~\ref{fig:pipeline}.
The key component of the framework, \ourBackbone, is introduced in~\ref{sec:pipeline}.
Then, the proposed \ours pre-training strategy is described in Sec.~\ref{sec:t-mae} and the corresponding windowed sparse cross-attention (WCA) module is elaborated in Sec.~\ref{sec:siamwca}.

\subsection{Preliminaries} \label{sec:preliminaries}
\boldparagraph{Pillar-based representation and sparse regional self-attention.} The pillar-based representation is an efficient representation introduced in PointPillars~\cite{lang2019pointpillars}. It divides 3D points into infinite-height voxels and computes pillar-wise features. These features can be treated as a pseudo-image in a bird's eye view.
On top of the pillar-based representation, SST~\cite{lFan_2022_SST} proposes a sparse regional self-attention (SRA) module to address challenges in applying ViT to sparse LiDAR points.
SST divides the pillars into non-overlapping windows and then applies self-attention within each window.
The receptive field of each pillar is expanded through a window shift operation.
This window partition and shift make the SRA module achieve a good balance of efficiency and accuracy.
\ours and many recent works~\cite{yang_gd-mae_2023, xu_mv-jar_2023, pSun_2022_SWFormer} are built upon the basic SRA block. 

\boldparagraph{Reconstruction pretext task and single-frame baseline.}
The original MAE~\cite{kHe_2021_MAE} randomly masks patches of an image and employs a ViT to reconstruct the image.
To perform this reconstruction pretext task in the LiDAR domain, we follow the concept of the state-of-the-art SSL method GD-MAE~\cite{yang_gd-mae_2023} which operates on a single frame.
The framework encompasses several key stages, \ie voxelization, masking, encoding, dense feature recovery, and reconstruction.
Through this pretext task, GD-MAE acquires valuable weights for the encoder, which are later utilized for downstream tasks.

\begin{figure*}[t] \centering
    \makebox[0.24\linewidth]{$t$}
    \makebox[0.24\linewidth]{${t-3} \ \sim \ t$}
    \hfill
    \makebox[0.20\linewidth]{$t$}
    \makebox[0.05\linewidth]{}{}
    \makebox[0.23\linewidth]{${t-3} \ \sim \ t$}
    \\
    \includegraphics[width=0.23\linewidth]{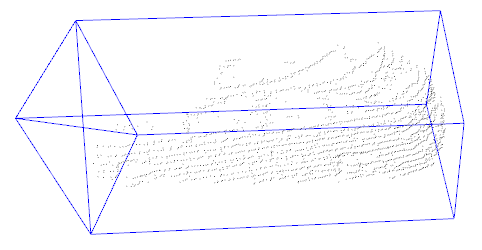} 
    \includegraphics[width=0.23\linewidth]{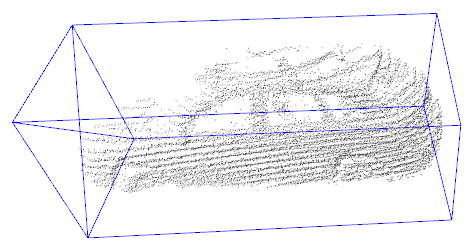} 
    \hfill
    \includegraphics[width=0.21\linewidth]{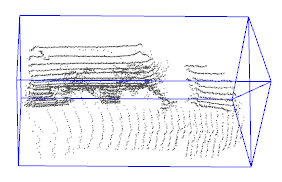}
    \includegraphics[width=0.285\linewidth]{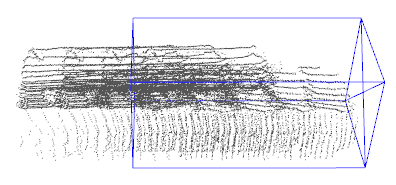} 
    \\
    \makebox[0.46\linewidth]{\small (a) A stationary vehicle.} 
    \hfill
    \makebox[0.46\linewidth]{\small (b) A moving vehicle.} 
    \caption{\textbf{Comparison between single- and four-frame concatenation.} While simple frame concatenation generally improves point density and detection rates, it can introduce spurious points in non-static scene parts that may degrade the detection performance. Since we combine consecutive frames via learned cross-attention, our approach is less affected by this problem. The blue bounding boxes indicate the ground truth for the current frame.} 
    \label{fig:veh_example}
\end{figure*}
\boldparagraph{Analysis.}
Up to this point, a common approach has been to employ a single-frame reconstruction pretext task for LiDAR points. However, incorporating historical frames poses a non-trivial challenge. One straightforward approach is to concatenate two point clouds after aligning them with ego-poses, similar to the three-frame variant of SST~\cite{lFan_2022_SST}. However, as shown in Fig.~\ref{fig:veh_example}, while concatenation helps to identify static objects effectively, it introduces challenges for detectors when dealing with moving objects. We provide quantitative results in Tab.~\ref{tab:two_frames}. To address this problem, we suggest learning from past data in a latent space.

\subsection{Framework Overview} \label{sec:pipeline}
\begin{figure*}[t]
  \centering
   \includegraphics[width=0.96\textwidth]{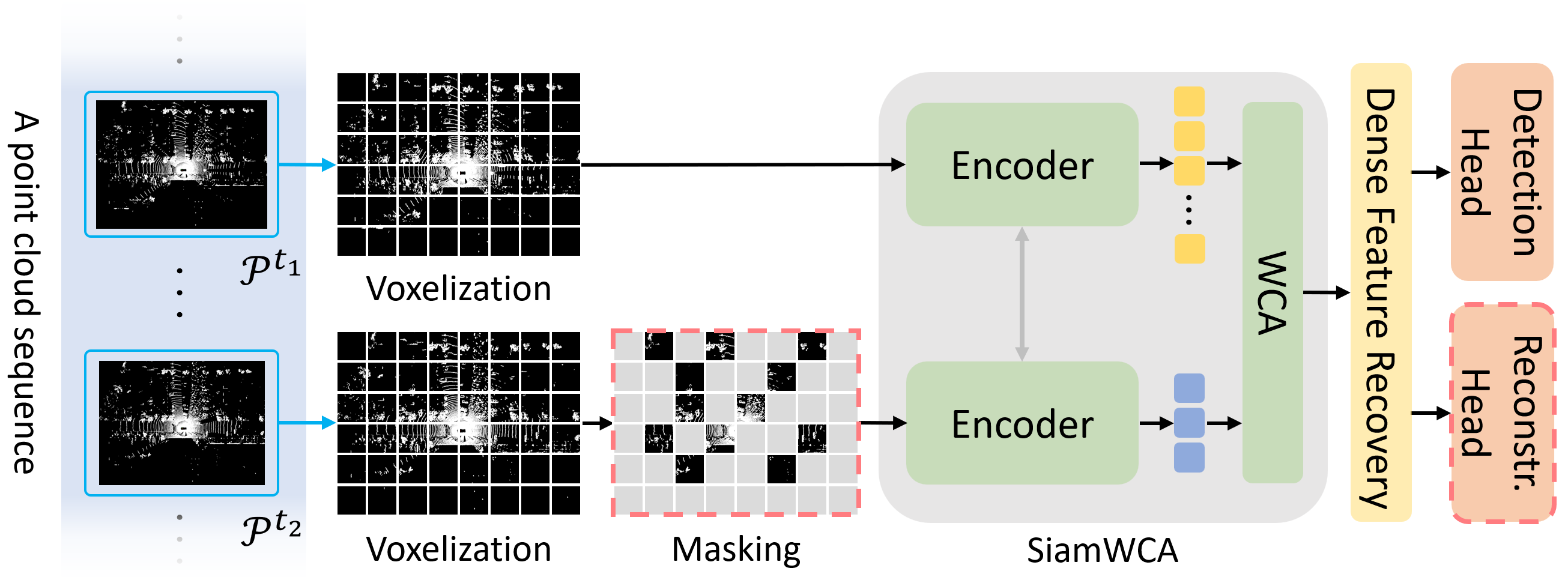}
   \caption{\textbf{Overview of our architecture and the proposed \ours pre-training.} Two frames are sampled from a sequence of point clouds and are voxelized. During pre-training, the current frame $\mathcal{P}^{t_2}$ undergoes an additional masking process. Note that the dashed boxes indicate operations for pre-training phase only. Next, voxel-wise tokens are computed by a Siamese encoder. The two-way gray arrow indicates weight sharing.
   The \ourFusion module takes as input the full tokens of the previous frame and the partial observation of the current frame and outputs enhanced tokens.
   The dense feature recovery places sparse tokens back to a dense feature map and convolves the map to fill empty locations.
   Subsequently, the feature map is either fed to a reconstruction head that recovers masked points, or to a detection head predicting bounding boxes.}
   \label{fig:pipeline}
\end{figure*}

As illustrated in Fig.~\ref{fig:pipeline}, the proposed framework compromises voxelization, encoding, windowed cross-attention, dense feature recovery and two separate heads for pre-training and detection, respectively. 
This subsection elaborates on the key components for supervised learning, \eg, training from scratch and fine-tuning, whereas SSL-related components are introduced in Sec.~\ref{sec:t-mae}.

\boldparagraph{Temporal batch-based sampling.}
Consider a sequence of point clouds as an ordered set of point clouds, denoted as $\mathbf{P} = \{\mathcal{P}^1, \mathcal{P}^2, \ldots, \mathcal{P}^t, \ldots, \mathcal{P}^T\}$. 
In this ordered set, $\mathcal{P}^t=\{(p^t_k)\}_{k=1}^K$ is a sweep of point cloud at time $t$, where $p_k$ is a 3D point $p_k \in \mathbb{R}^3$ and $K$ indicates the number of points.
In our setting, two frames are needed to feed the network.
\if\RefAppendix1
Due to the high sampling frequency of the LiDAR system, two consecutive frames usually contain redundant duplicate information, which is verified in Sec.~\ref{sec:exp_temporal_gap}.
\else
Due to the high sampling frequency of the LiDAR system, two consecutive frames usually contain redundant duplicate information, which is verified in Appendix Sec. S3.
\fi
Conversely, when there is a substantial time gap between two frames, their overlap may be limited, making the historical information less useful. 
Therefore, we follow TARL~\cite{nunes_tarl} to introduce the concept of a temporal batch so that the interval of two frames is constrained to an appropriate range.
Specifically, we sample a batch of consecutive frames $\mathcal{B}^t = \{\mathcal{P}^{t+1}, \mathcal{P}^{t+2}, \ldots, \mathcal{P}^{t+n}\}$ from a sequence $\mathcal{P}$.
Then, two frames, $\mathcal{P}^{t_1}$ and $\mathcal{P}^{t_2}$  are sampled from the first and last one-third of the batch, namely, $t_1 \in \{t+1, t+2, \ldots, \left\lfloor t+\frac{n}{3} \right\rfloor \}$ and $t_2 \in \{ \left\lceil t+\frac{2n+1}{3} \right\rceil , \ldots, n-1, n\}$.

\boldparagraph{Alignment and voxelization.}
Given two frames, the previous frame $\mathcal{P}^{t_1}$ is transformed to the coordinate system of the current frame $\mathcal{P}^{t_2}$ by their ego-poses.
The pose information is available in most datasets~\cite{pSun_2020_Waymo, nuscenes, jMao_2021_ONCE} or easily obtained by means of GPS/IMU, odometry approaches~\cite{xChen_2019_SuMa}, structure from motion (SfM) algorithms~\cite{schoenberger2016sfm, schoenberger2016mvs}, or SLAM systems~\cite{pDellenbach_2022_CT-ICP}.
Next, each point cloud is divided into discrete voxels.
Two linear layers map point coordinates to high-dimensional features and a voxel-wise representation is obtained via voxel-wise average pooling.
The voxels can be regarded as pillars, owing to the infinite height.

\boldparagraph{A Siamese encoder and windowed cross-attention (SiamWCA).}
A Siamese encoder~\cite{SiameseNetwork1993} is a two-branch network where both branches share the same configurations and weights.
In this work, it is utilized to encode the pillar-wise representations of both frames to sparse tokens.
These tokens serve as input to the \ourFusion module which facilitates the interaction between historical and current tokens. 
The SiamWCA is elaborated in Sec.~\ref{sec:siamwca}.

\boldparagraph{Dense feature recovery and detection head.}
Once the current tokens are augmented with historical information, these sparse tokens are reverted to the x-y plane to form a dense feature map while vacant pillars are filled with zeros.
Since LiDAR points only occur on object surfaces, the object centers typically locate at empty space, leading to inaccurate detection.
We follow GD-MAE~\cite{yang_gd-mae_2023} to attach four dense convolutional layers, which spread the feature from occupied pillars to vacant regions.
For the detection head, we adopt a center-based head and use the same target assignment strategy as CenterPoint~\cite{tYin_2021_CenterPoint}.

\subsection{Temporal Masked Autoencoder Pre-training} \label{sec:t-mae}
Up to this point, a two-frame framework for object detection has been set up. 
However, the transformer-based architecture is data-hungry.
Inspired by the SiamMAE~\cite{aGupta_2023_SiameseMAE} used in video understanding, we develop \emph{Temporal Masked Auto-Encoders} (T-MAE) for self-supervised learning on LiDAR points. 
As shown in Fig.~\ref{fig:pipeline}, the core idea is to reconstruct the present frame based on a full observation of the historical frame and a partial observation of the current frame. 
In this way, the network is compelled to learn a powerful sparse representation as well as the capacity to effectively model motion. 
After the pre-training, the weights of the \ourBackbone backbone are retained for the downstream tasks.
In the subsequent paragraphs, we elaborate on these steps.

\boldparagraph{Masking.}
As shown in Fig.~\ref{fig:pipeline}, on top of the proposed \ourBackbone backbone, an additional step, namely masking, is added between the voxelization and the encoder for the current frame.
Specifically, masking is applied in a pillar-wise manner for a high ratio of the occupied pillars, \eg, 75\%.
The remaining pillars are subsequently input into the encoder.
The encoders for the two frames share weights to ensure that the features are constrained in the same latent space.
Due to masking, the number of tokens for the current frame decreases significantly, leading to much fewer valid windows and thus accelerating WCA during pre-training.
The dense feature recovery performs exactly the same as in Sec.~\ref{sec:pipeline} and outputs a dense feature map.

\boldparagraph{Reconstruction head.}
Given the dense feature map, the reconstruction head retrieves the feature of masked pillars by their spatial location.
With the pillar-wise features, the head reconstructs the relative coordinates of points, where the number of output points per pillar is set as $K^{O}$.
Eventually, the Chamfer distance is computed as the loss function between the reconstructed points and the ground-truth points. 
Note that the number of points per pillar varies drastically. 
Thus, a fixed number of points $K^{GT}$ are randomly sampled as the target for reconstruction.
In conclusion, T-MAE masks the voxelized tokens of the current frame and compels the network to reconstruct the current frame with the full observation of the historic frame as a reference, which encourages the WCA module to build up correspondence between frames.
\if\RefAppendix1
Note that, unlike the single-frame baseline where only the encoder is reused for downstream tasks, the weights of \ourBackbone are fully retained for downstream tasks, which is proved to be effective as shown in Tab.~\ref{tab:wca_random_init}.
\else
Note that, unlike the single-frame baseline where only the encoder is reused for downstream tasks, the weights of \ourBackbone are fully retained for downstream tasks, which is proved to be effective as shown in Appendix Tab. S10.
\fi
Therefore, the capability for building up correspondence is also retained.

\subsection{Siamese Encoder and Windowed Cross-Attention (SiamWCA)} \label{sec:siamwca}
\begin{figure*}[t]
  \centering
   \includegraphics[width=0.98\textwidth]{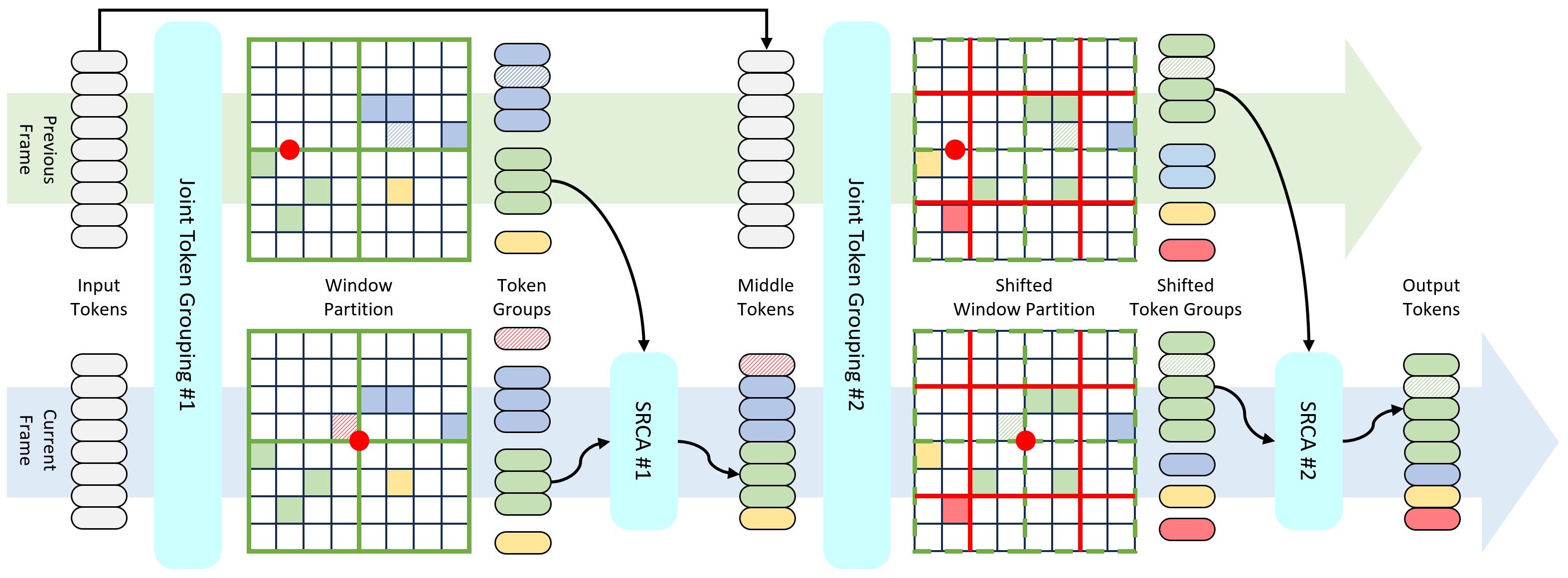}
   \caption{\textbf{Windowed sparse cross-attention (WCA).} Given the input tokens from both $\mathcal{P}^{t_1}$ and $\mathcal{P}^{t_2}$, a joint token grouping is performed to obtain a window partition. A sparse regional cross-attention (SRCA) is performed independently in each window to integrate the historical information to the middle tokens of the current frame. In other words, the tokens from two frames but with the same colors are attending to each other. For simplicity, the information flow is only depicted for the green tokens. After the second joint token grouping, the cross-attention are performed once more with the shifted window partition. The red dot (\protect\includegraphics[height=1.5ex]{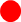}) indicates the ego-vehicle driving towards the right. The box with diagonal stripes (\protect\includegraphics[height=1.5ex]{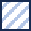}) represents an object, \eg, a vehicle, moving towards the left. Best viewed in color and high-resolution.}
   \label{fig:WCA}
\end{figure*}

\boldparagraph{Siamese encoder.}
Given the pillar-wise representations of a pair of frames, we explore an asymmetric network and a Siamese encoder for feature encoding.
The \textit{asymmetric network} consists of two branches with the same architecture but the branch for $\mathcal{P}^{t_1}$ is modified by reducing the number of channels by half, as depicted in Fig.~\ref{fig:arch_alternatives} (a).
A \textit{Siamese encoder} is a symmetric network with two subnetworks sharing weights.
It is widely used to compute similarity in latent space for tracking~\cite{lBertinetto_2016_SiamFC, aHe_2018_SA-Siam, rTao_2016_SINT, qGuo_2017_DSiam} and contrastive learning~\cite{xChen_2021_SimSiam, xPeng_2022_ContrastiveCrop,zChen_2023_SiameseDETR}.
For the Siamese encoder, we investigate two weight updating strategies, namely accumulation and SimSiam-style~\cite{xChen_2021_SimSiam}.
Accumulation indicates the backward gradients of the two encoders are accumulated.
SimSiam, standing for the simple Siamese pre-training strategy~\cite{xChen_2021_SimSiam}, indicates the encoder for $\mathcal{P}^{t_1}$ is detached from the computational graph, as depicted in Fig.~\ref{fig:arch_alternatives} (b). 
Thus the weight of this encoder is not updated by gradient propagation but by copying the weights from the encoder for $\mathcal{P}^{t_2}$.

\boldparagraph{Windowed cross-attention (WCA).}
When provided with input tokens from two frames, there are several methods for interaction.
One intuitive approach is to apply a standard self-attention layer to the concatenation of all tokens from both frames. 
This approach significantly increases GPU memory requirements because it doubles the number of input tokens. 
The vanilla Transformer block~\cite{aVaswani_2017_Transformer}, which consists of a cross-attention layer and a self-attention layer can also be adopted and a single cross-attention layer is feasible as well.
However, all these conventional global attentions are not affordable in 3D space due to the expensive computational overhead imposed by the unavoidable high resolution. 
For instance, the input for our Siamese encoder can be $\mathbb{R}^{(468 \times 468) \times 128}$, whereas it is $\mathbb{R}^{(14 \times 14) \times 384}$ for a ViT-B/16 due to the patch embedding.
Therefore, a window-based implementation is crucial for efficiency.
In brief, the \ourFusion module divides the 3D space into non-overlapping windows and then performs cross-attention within the windows.
Figure~\ref{fig:WCA} intuitively illustrates this process with a diagram that simulates a scene from a bird's eye view.
The key components are elaborated in the following paragraphs.

\boldparagraph{$\triangleright$ Joint token grouping}
partitions the 3D space into non-overlapping windows and subsequently allocates the token to the corresponding window based on its spatial location.
The previous frame has been aligned with the current frame and thus the window partition is unified for both frames, which allows the following attention mechanism to perform within each window.
As shown in Fig.~\ref{fig:WCA}, all tokens within the same physical window share the same colour, indicating that they are assigned to the same group for mutual attention.

\boldparagraph{$\triangleright$ Sparse Regional Cross-Attention (SRCA)} is essentially a cross-attention layer where the query comes from $\mathcal{P}^{t_2}$ and the key-value comes from $\mathcal{P}^{t_1}$.
For clarity, we consider a single window as an example.
Given two groups of tokens $\mathcal{F}^{t_1}, \mathcal{F}^{t_2}$ from two frames and their corresponding spatial coordinates $\mathcal{I}^{t_1}, \mathcal{I}^{t_2}$, the cross-attention is performed as follows:
\begin{align}			    
  \hat{\mathcal{F}}^{t_2} &= \mathbf{MCA} \left ( \mathcal{F}^{t_2} + \mathbf{PE}(\mathcal{I}^{t_2}), \mathcal{F}^{t_1} + \mathbf{PE}(\mathcal{I}^{t_1}), \mathcal{F}^{t_1} \right ) \\
  \widetilde{\mathcal{F}}^{t_2} &= \mathbf{LN}\left ( \mathbf{MLP}\big(\mathbf{LN}(\hat{\mathcal{F}}^{t_2})\big) + \hat{\mathcal{F}}^{t_2} \right ) + \mathcal{F}^{t_2}
  \label{eq:method_srca}
\end{align}
where $\mathbf{MCA}(\mathcal{Q}, \mathcal{K}, \mathcal{V})$ indicates a classical Multi-head Cross-Attention, $\mathbf{PE}(\cdot)$ represents the absolute positional encoding function used in~\cite{nCarion_2020_DETR}, and $\mathbf{LN}(\cdot)$ stands for Layer Normalization.
Note that, if a window is empty in the previous frame, the tokens of this window in the current frame will remain unchanged, \ie $ \widetilde{\mathcal{F}}^{t_2} = \mathcal{F}^{t_2} $.
While the core concept of WCA is identical to cross-attention, this windowed implementation significantly scales down the computational complexity especially when working with sparse pillars.

\boldparagraph{$\triangleright$ Repeated operations with window shift.} 
Given the middle tokens, the window partition is shifted by half of the window size.
Next, the tokens are re-grouped by the joint token grouping, as illustrated in ``Shifted Window Partition'' of Fig.~\ref{fig:WCA}.
After that, the SRCA is performed once more.
Note that, while performing the SRCA \#2, the tokens $\widetilde{\mathcal{F}}^{t_2}$ of $\mathcal{P}^{t_2}$ are updated by SRCA \#1 while the tokens $\mathcal{F}^{t_1}$ remain unchanged because a cross-attention operation does not update key $\mathcal{K}$ and value $\mathcal{V}$.
With this window shift and SRCA \#2, a token interacts with more tokens.
For instance, the yellow token in the first window partition only attaches itself in the previous frame via SRCA \#1 but interacts with more tokens in SRCA \#2.

\section{Experiments}
\label{sec:exp}

\subsection{Dataset and Implementation} \label{sec:implementation}
Experiments are conducted on the following two datasets.

\boldparagraph{Waymo Open dataset~\cite{pSun_2020_Waymo}} is a large-scale autonomous driving dataset with LiDAR points. 
For 3D detection, an evaluation protocol is provided for calculating the average precision (AP) and the average precision weighted by heading (APH). 
Moreover, the evaluation includes two difficulty levels wherein bounding boxes containing over 5 points are regarded as Level 1, while Level 2 indicates all bounding boxes. 
We adopt mean AP and mean APH at Level 2 as the main evaluation metrics.

\boldparagraph{ONCE dataset~\cite{jMao_2021_ONCE}} consists of 581 sequences of varying length.
6, 4 and 10 sequences are selected for training, validation and test sets, respectively, and manually annotated.
The remaining data is not annotated and adopted for pre-training in our experiments.
The official evaluation metrics are AP calculated for each category and range.

\boldparagraph{Implementation Details.} 
We implement our approach based on the codebase of OpenPCDet\footnote{\url{https://github.com/open-mmlab/OpenPCDet}}.
We adopt the sparse pyramid transformer as our encoder and keep the configuration consistent with its official implementation of GD-MAE~\cite{yang_gd-mae_2023}.
Three data augmentation techniques, \ie random flipping, scaling, and rotation, are applied to both frames during pre-training and finetuning.
In addition, following MV-JAR~\cite{xu_mv-jar_2023} and GD-MAE~\cite{yang_gd-mae_2023}, a copy-n-paste augmentation~\cite{yYan_2018_SECOND} is also employed to slightly address the issue of class imbalance during finetuning.
Note that identical data augmentations are applied to a pair of two frames.  
Further details are in the supplementary material.

\subsection{Main Results}
We present the comparison with SOTA methods on two datasets:
For the ONCE dataset~\cite{jMao_2021_ONCE}, we pre-train our \ourBackbone with the \textit{raw\_large} split and fine-tune it with the annotated training set.
For the Waymo dataset~\cite{pSun_2020_Waymo}, we pre-train \ourBackbone with the entire training set.
Then, following MV-JAR~\cite{xu_mv-jar_2023}, we finetune our approach with four portions of labeled data, namely 5\%, 10\%, 20\% and 100\%.
Note that, a strong counterpart, GD-MAE~\cite{yang_gd-mae_2023}, has not been evaluated in this setting and thus we re-train it for a more comprehensive comparison.

\begin{table*}[t]
\caption{ \textbf{Comparison with SSL methods on the Waymo validation set~\cite{pSun_2020_Waymo}.} Random initialization denotes training from scratch. $\dag$ represents duplicating the current frame as input during inference. $^*$ and $^{**}$ indicate reproduced by us and taken from~\cite{yang_gd-mae_2023}, respectively. Results for other methods are taken from MV-JAR~\cite{xu_mv-jar_2023} or the survey~\cite{bFei_2023_PCSSL_Survey}. Best results are highlighted as \colorbox{colorFst}{\bf first}, \colorbox{colorSnd}{second}, and \colorbox{colorTrd}{third}. Differences between \ours pre-training and random initialization are highlighted in \red{red}.
}
\centering
\scalebox{0.71}{\tablestyle{6pt}{1.0}
\begin{tabular}{c|l|ll|cccccc}
\toprule
\multirow{2}{*}{\makecell{Data\\Amount}} & \multirow{2}{*}{Initialization} & \multicolumn{2}{c|}{Overall} & \multicolumn{2}{c}{Vehicle} & \multicolumn{2}{c}{Pedestrian} & \multicolumn{2}{c}{Cyclist} \\ \cmidrule(lr){3-4} \cmidrule(lr){5-6} \cmidrule(lr){7-8} \cmidrule(lr){9-10}
 &  & mAP & mAPH & mAP & mAPH & mAP & mAPH & mAP & mAPH \\ \midrule
\multirow{7}{*}{5\%} & Random                           & 43.68 & 40.29 & 54.05 & 53.50 & 53.45 & 44.76 & 23.54 & 22.61 \\
 & PointContrast~\cite{sXie_2020_PointContrast}         & 45.32 & 41.30 & 52.12 & 51.61 & 53.68 & 43.22 & 30.16 & 29.09 \\
 & ProposalContrast~\cite{jYin_2022_ProposalContrast}   & 46.62 & 42.58 & 52.67 & 52.19 & 54.31 & 43.82 & 32.87 & 31.72 \\
 & MV-JAR~\cite{xu_mv-jar_2023}                         & \rd 50.52 & \rd 46.68 & \rd 56.47 & \rd 56.01 & \rd 57.65 & \rd 47.69 & \nd 37.44 & \nd  36.33 \\
 & GD-MAE~\cite{yang_gd-mae_2023}$^*$                   & 48.23 & 44.56 & 56.34 & 55.76 & 55.62 & 46.22 & 32.72 & 31.69 \\
 & T-MAE$^\dag$                                         & \nd 50.89 & \nd 47.22 & \nd 57.06 & \nd 56.05 & \nd 58.95 & \nd 52.62 & \rd 36.64 & \rd 32.99  \\
 & \textbf{\ours (Ours)}                                & \fs 51.47\up{7.79} & \fs 49.46\up{9.17} & \fs 57.13 & \fs 56.63 & \fs 59.69 & \fs 55.28 & \fs 37.61 & \fs 36.48 \\ 
 \midrule
\multirow{7}{*}{10\%} & Random                          & 56.05 & 53.13 & \rd 59.78 & \rd 59.27 & 60.08 & 53.04 & 48.28 & 47.08 \\
 & PointContrast~\cite{sXie_2020_PointContrast}         & 53.69 & 49.94 & 54.76 & 54.30 & 59.75 & 50.12 & 46.57 & 45.39 \\
 & ProposalContrast~\cite{jYin_2022_ProposalContrast}   & 53.89 & 50.13 & 55.18 & 54.71 & 60.01 & 50.39 & 46.48 & 45.28 \\
 & MV-JAR~\cite{xu_mv-jar_2023}                         & 57.44 & 54.06 & 58.43 & 58.00 & \nd 63.28 & \rd 54.66 & 50.63 & 49.52 \\
 & GD-MAE~\cite{yang_gd-mae_2023}$^*$                   & \rd 57.67 & \rd 54.31 & 59.72 & 59.19 & 60.43 & 52.21 & \nd 52.85 & \nd 51.52  \\
 & T-MAE$^\dag$                                         & \nd 58.52 & \nd 55.59 & \nd 60.26 & \nd 59.75 & \rd 62.89 & \nd 55.85 & \rd 52.43 & \rd 51.16 \\
 & \textbf{\ours (Ours)}                                & \fs 59.93\up{3.88} & \fs 57.99\up{4.86} & \fs 60.27 & \fs 59.77 & \fs 65.23 & \fs 61.10 & \fs 54.29 & \fs 53.09 \\
 \midrule
\multirow{7}{*}{20\%} & Random                          & 60.21 & 57.61 & 61.58 & 61.08 & 64.63 & 58.41 & 54.42 & 53.33 \\
 & PointContrast~\cite{sXie_2020_PointContrast}         & 59.35 & 55.78 & 58.64 & 58.18 & 64.39 & 55.43 & 55.02 & 53.73 \\
 & ProposalContrast~\cite{jYin_2022_ProposalContrast}   & 59.52 & 55.91 & 58.69 & 58.22 & 64.53 & 55.45 & 55.36 & 54.07 \\
 & MV-JAR~\cite{xu_mv-jar_2023}                         & 62.28 & \rd 59.15 & 61.88 & 61.45 & \rd 66.98 & \rd 59.02 & \rd 57.98 & \rd 57.00 \\
 & GD-MAE~\cite{yang_gd-mae_2023}$^*$                   & \rd 62.32 & 59.09 & \nd 62.27 & \nd 61.79 & 66.12 & 58.06 & \nd 58.57 & \nd 57.42 \\
 & T-MAE$^\dag$                                         & \nd 62.37 & \nd 60.17 & \rd 62.19 & \rd 61.72 & \nd 67.18 & \nd 62.18 & 57.74 & 56.59 \\
 & \textbf{\ours (Ours)}                                & \fs 63.52\up{3.31} & \fs 61.80\up{4.19} & \fs 63.10 & \fs 62.59 & \fs 68.23 & \fs 64.66 & \fs 59.23 & \fs 58.15 \\
 \midrule
\multirow{9}{*}{100\%} & Random                         & \rd 71.30 & \nd 69.13 & \rd 69.05 & \rd 68.62 & \rd 73.77 & \nd 68.80 & \nd 71.09 & \nd 69.97  \\
 & GCC-3D~\cite{hLiang_2021_ExploringGeometryawareContrast} & 65.29 & 62.79 & 63.97 & 63.47 & 64.23 & 58.47 & 67.88 & 66.44 \\
 & BEV-MAE~\cite{lin_bev-mae_2022}                      & 66.92 & 64.45 & 64.78 & 64.29 & 66.25 & 60.53 & 69.73 & 68.52  \\
 & PointContrast~\cite{sXie_2020_PointContrast}         & 68.06 & 64.84 & 64.24 & 63.82 & 71.92 & 63.81 & 68.03 & 66.89 \\
 & ProposalContrast~\cite{jYin_2022_ProposalContrast}   & 68.17 & 65.01 & 64.42 & 64.00 & 71.94 & 63.94 & 68.16 & 67.10 \\
 & MV-JAR~\cite{xu_mv-jar_2023}                         & 69.16 & 66.20 & 65.52 & 65.12 & 72.77 & 65.28 & 69.19 & 68.20 \\ 
 & GD-MAE~\cite{yang_gd-mae_2023}$^{**}$                & 70.62 & 67.64 & 68.72 & 68.29 & 72.84 & 65.47 & 70.30 & 69.16  \\
 & T-MAE$^\dag$                                         & \nd 71.56 & \rd 69.00 & \nd 69.39 & \nd 68.95 & \nd 74.42 & \rd 68.43 & \rd 70.86 & \rd 69.61 \\
 & \textbf{\ours (Ours)}                                & \fs 72.30\up{1.00} & \fs 70.52\up{1.39} & \fs 69.34 & \fs 68.89 & \fs 75.79 & \fs 72.01 & \fs 71.78 & \fs 70.65  \\
\bottomrule
\end{tabular}
}
\label{tab:waymo_main}
\end{table*}

\boldparagraph{The impact of the \ours pre-training.}
The bottom block in Tab.~\ref{tab:waymo_main} shows that the randomly initialized \ourBackbone (denoted as \textit{Random}) performs better than any other models that are initialized with a different pre-training strategy (\ie 69.13 \textit{v.s.} 67.64), suggesting that \ourBackbone is a powerful backbone capable of learning temporal modeling when provided with sufficient annotated data.
However, its performance drops significantly when finetuning data is limited (\eg, 40.29 \textit{v.s.} 46.68 at 5\% level), indicating a strong demand for annotated data.
As a comparison, the proposed \ours consistently enhances \ourBackbone compared to random initialization.
Moreover, as the labeled data shrinks from 100\% to 5\%, the impact of the \ours pre-training becomes more pronounced, namely increasing from 1.39 to 9.17, suggesting that the pre-training approach learns a powerful representation and alleviates the demand for annotated data.

\begin{figure*}[t] \centering
    \includegraphics[width=0.49\linewidth]{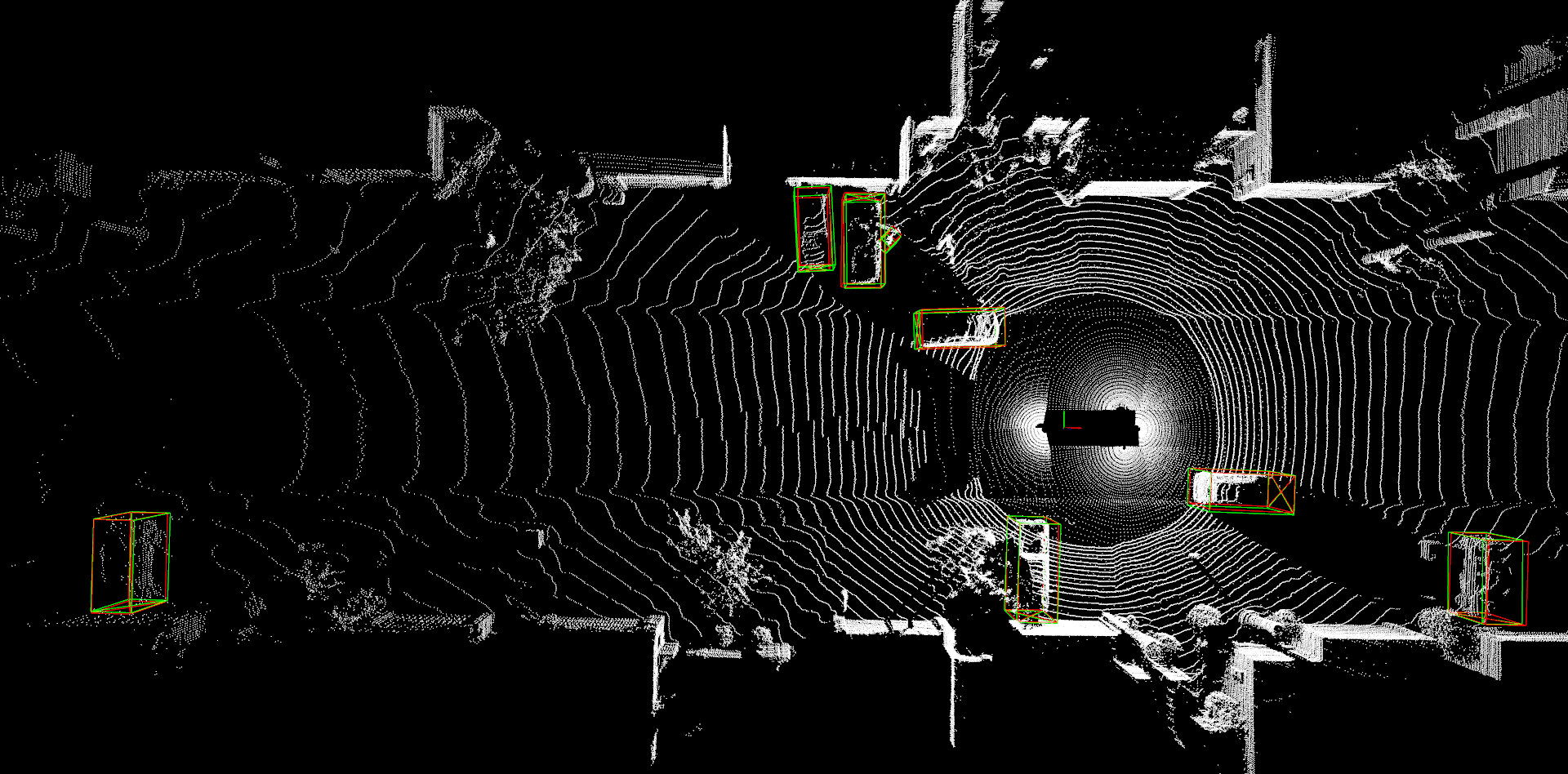} 
    \includegraphics[width=0.49\linewidth]{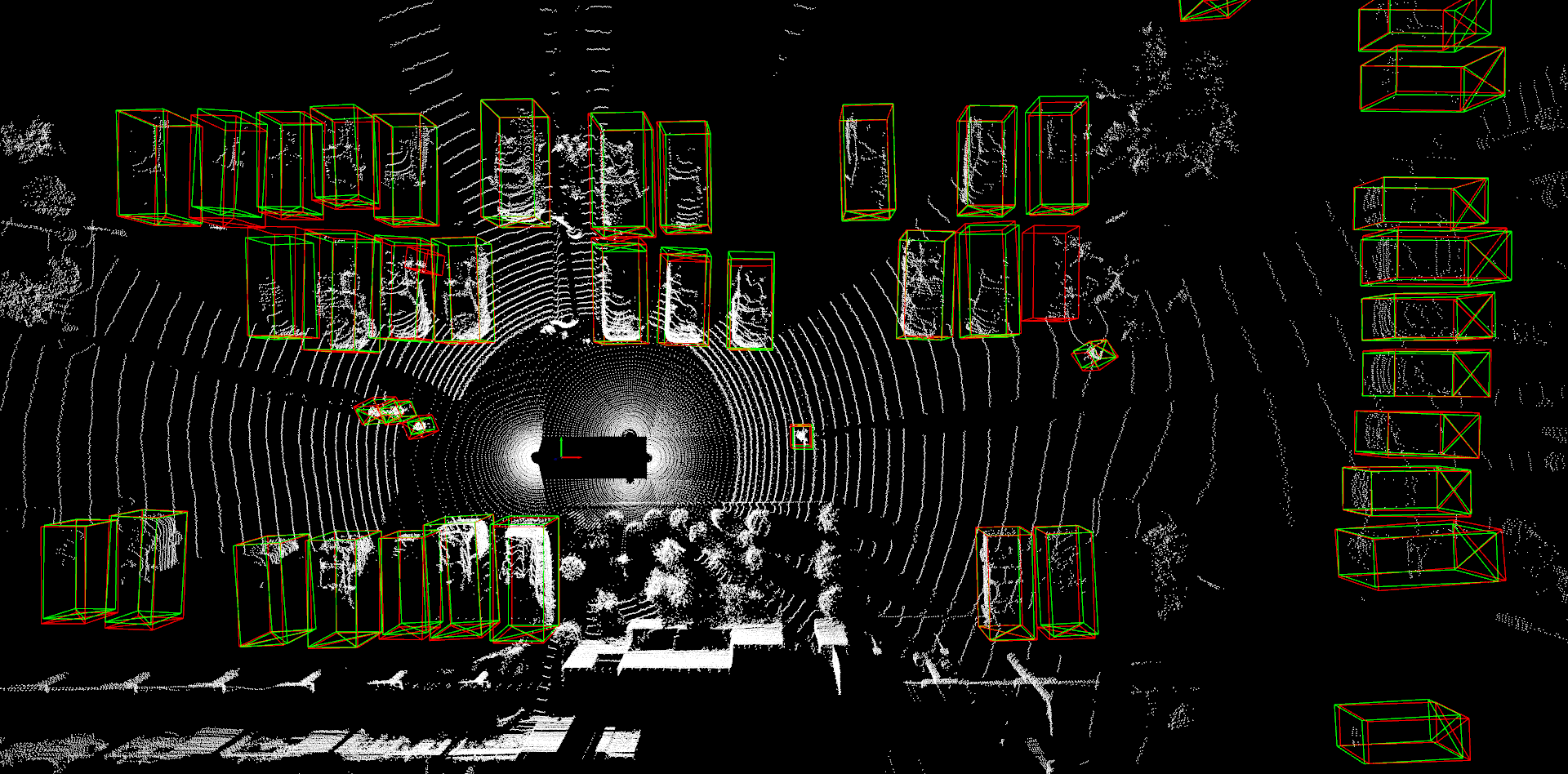}
    \caption{\textbf{Qualitative results.} We depict ground truth and predictions as boxes colored in red and green for two exemplary scenes from the Waymo dataset~\cite{pSun_2020_Waymo}.} 
    \label{fig:qua_results}
\end{figure*}

\boldparagraph{Comparison with SOTA methods.}
We aim to leverage the temporal information between two adjacent frames, which is often overlooked by other methods.
This absence makes it challenging to compare our method with others in the same setting.
Therefore, we implement a test-time single-frame baseline (denoted as T-MAE$^\dag$) by replicating the same frame and inputting them into our pre-trained model during evaluation.
Table~\ref{tab:waymo_main} shows that \ours with identical frames outperforms SOTA counterparts.
Moreover, \ours with adjacent frames achieves new SOTA at all levels in terms of overall and class-specific metrics.
Notably, thanks to the temporal modeling ability, \ours significantly outperforms other methods in terms of L2 mAPH for pedestrians. For instance, with 5\% labeled data, \ours achieve better mAPH (\ie 55.28) than any other method using 10\% labeled data.
This metric indicates better direction detection for pedestrians, which could benefit downstream applications, \eg, pedestrian intention prediction.

\if\RefAppendix1
To reduce performance variance on 5\% and 10\% splits, we also report \ours performance in the Appendix (see Sec.~\ref{sec:additional_waymo}) when it is finetuned with two more 5\% and 10\% splits and another 20\% split.
\ours consistently outperforms other methods, indicating the efficacy of \ours.
Since there are no existing multi-frame SSL methods, we also compare our approach with robust non-SSL baselines that utilize multi-frame as inputs in the Appendix (see Tab.~\ref{tab:multiframe}).
\else
To reduce performance variance on 5\% and 10\% splits, we also report \ours performance in the Appendix (see Sec. S4) when it is finetuned with two more 5\% and 10\% splits and another 20\% split.
\ours consistently outperforms other methods, indicating the efficacy of \ours.
Since there are no existing multi-frame SSL methods, we also compare our approach with robust non-SSL baselines that utilize multi-frame as inputs in the Appendix (see Tab. S9).
\fi

To assess the generalization capabilities of our method, we also conducted experiments on the ONCE dataset~\cite{jMao_2021_ONCE}.
As shown in Tab.~\ref{table:once_full}, \ours outperforms other methods in most metrics, indicating its superiority.
Moreover, the substantial improvement for pedestrians generalizes to this new dataset, indicating the dominance of \ours in pedestrian detection.

Figure~\ref{fig:qua_results} shows two exemplary qualitative results from the Waymo dataset and more qualitative results are presented in the Appendix.

\begin{table*}[t]
    \caption{Performance comparisons on the validation split of the ONCE dataset~\cite{jMao_2021_ONCE}. Pt. indicates the model is initialized with pre-trained weights. Results for other methods are taken from GD-MAE~\cite{yang_gd-mae_2023}.}
    \centering
    \scalebox{0.5}{\tablestyle{6pt}{1.0}
    \begin{tabular}{l|c|l|cccc|cccc|cccc}
    \toprule
    \multirow{2}{*}{Methods} & \multirow{2}{*}{Pt.} &
    \multirow{2}{*}{mAP} &
    \multicolumn{4}{c|}{Vehicle} & \multicolumn{4}{c|}{Pedestrian} & \multicolumn{4}{c}{Cyclist} \\
     &&& Overall & 0-30m & 30-50m & 50m-Inf & Overall & 0-30m & 30-50m & 50m-Inf & Overall & 0-30m & 30-50m & 50m-Inf \\
    \midrule
    PV-RCNN~\cite{sShi_2020_PV-RCNN} & \redx                & 53.55 & \nd 77.77 & \fs 89.39 & \nd 72.55 & \fs 58.64 & 23.50 & 25.61 & 22.84 & 17.27 & 59.37 & 71.66 & 52.58 & 36.17 \\
    IA-SSD~\cite{zhang2022iassd} & \redx                    & 57.43 & 70.30 & 83.01 & 62.84 & 47.01 & 39.82 & 47.45 & 32.75 & 18.99 & 62.17 & 73.78 & 56.31 & 39.53 \\
    CenterPoint-Pillar~\cite{tYin_2021_CenterPoint} & \redx & 59.07 & 74.10 & 85.23 & 69.22 & 53.14 & 40.94 & 48.43 & 34.72 & 20.09 & 62.17 & 73.70 & 56.05 & 40.19 \\
    CenterPoint-Voxel~\cite{tYin_2021_CenterPoint} & \redx  & 60.05 & 66.79 & 80.10 & 59.55 & 43.39 & 49.90 & 56.24 & \nd 42.61 & \fs 26.27 & 63.45 & 74.28 & 57.94 & 41.48 \\
    \midrule
    SECOND~\cite{yYan_2018_SECOND} & \redx                  & 51.89 & 71.19 & 84.04 & 63.02 & 47.25 & 26.44 & 29.33 & 24.05 & 18.05 & 58.04 & 69.96 & 52.43 & 34.61 \\
    w/ BYOL~\cite{jbGrill_2020_BYOL} & \greencheck                & 51.63 & 71.32 & 83.59 & 64.89 & 50.27 & 25.02 & 27.06 & 22.96 & 17.04 & 58.56 & 70.18 & 52.74 & 36.32 \\
    w/ PointContrast~\cite{sXie_2020_PointContrast} & \greencheck & 53.59\up{1.70} & 71.87 & 86.93 & 62.85 & 48.65 & 28.03 & 33.07 & 25.91 & 14.44 & 60.88 & 71.12 & 55.77 & 36.78 \\
    w/ DeepCluster~\cite{DeepCluster} & \greencheck               & 53.72\up{1.83} & 72.89 & 83.52 & 67.09 & 50.38 & 30.32 & 34.76 & 26.43 & 18.33 & 57.94 & 69.18 & 52.42 & 34.36 \\
    \midrule
    SPT~\cite{yang_gd-mae_2023} & \redx                     & 62.62 & 75.64 & 87.21 & 70.10 & 53.21 & 45.92 & 54.78 & \rd 37.84 & 22.56 & 66.30 & 78.12 & 60.52 & 42.05 \\
    w/ GD-MAE~\cite{yang_gd-mae_2023} & \greencheck               & \nd 64.92\up{2.30} & \rd 76.79 & \rd 88.01 & \rd 71.70 & \rd 55.60 & \nd 48.84 & \nd 58.70 & 37.30 & \nd 25.72 & \nd 69.14 & \nd 80.29 & \nd 64.58 & \nd 45.14 \\
    \midrule
    \textbf{SiamWCA (Ours)}        & \redx                  & \rd 63.71 & 76.47 & 87.63 & 71.59 & 55.16 & \rd 47.27 & \rd 57.57 & 36.99 & 21.79 & \rd 67.40 & \rd 78.39 & \rd 62.78 & \rd 43.90    \\
    \textbf{w/ \ours (Ours)}   & \greencheck                      & \fs 67.00\up{3.29} & \fs 78.35 & \nd 88.45 & \fs 73.05 & \nd 57.16 & \fs 52.57 & \fs 62.66 & \fs 44.18 & \rd 25.29 & \fs 70.09 & \fs 81.14 & \fs 65.33 & \fs 46.48 \\
    \bottomrule
    \end{tabular}
    }
    \label{table:once_full}
\end{table*}

\subsection{Ablation Study} \label{sec:ablation}

We perform ablation studies on the Waymo dataset~\cite{pSun_2020_Waymo} to justify design choices and hyperparameters. 
For cost-effective experiments, the split 0 with 5\% data of the data-efficient benchmark~\cite{yang_gd-mae_2023} is used to finetune the model by default.

\boldparagraph{Is a delicate fusion module necessary?}
\begin{table}[tb]
\caption{\textbf{Two frames comparison.} Two consecutive frames are merged and input into the GD-MAE~\cite{yang_gd-mae_2023} as an enhanced baseline.}
\centering
\tablestyle{3pt}{1.0}
\scriptsize
\begin{tabular}{lcccccccccc}
\toprule
\multirow{2}{*}{Method}  & \multicolumn{2}{c}{Frame Input} & \multicolumn{2}{c}{L2 Overall} & \multicolumn{2}{c}{Vehicle} & \multicolumn{2}{c}{Pedestrian} & \multicolumn{2}{c}{Cyclist} \\ 
\cmidrule(lr){2-3} \cmidrule(lr){4-5} \cmidrule(lr){6-7} \cmidrule(lr){8-9} \cmidrule(lr){10-11}
                &  Previous  & Current & mAP   & mAPH    & AP   & APH    & AP    & APH     & AP     & APH    \\ \midrule
GD-MAE          & -  & 0        & \nd 48.23    & \nd 44.56   & \rd 56.34   & \rd 55.76  & \rd 55.62    & \rd 46.22   & \nd 32.72  & \nd 31.69  \\
GD-MAE          & -  & \{-1, 0\}    & \rd 47.35    & \rd 43.69   & \fs 57.04   & \fs 56.48  & \nd 57.76    & \nd 48.48   & \rd 27.26  & \rd 26.12  \\
\ours (Ours)    & -1 & 0        & \fs 49.45 & \fs 46.78 & \nd 56.56 & \nd 56.02 & \fs 57.96 & \fs 51.94 & \fs 33.82 & \fs 32.37 \\ 
\bottomrule
\end{tabular}
\label{tab:two_frames}
\end{table}

To compare with a model taking two frames as input, we modified the input of GD-MAE~\cite{yang_gd-mae_2023} by concatenating all points of two consecutive frames. 
We pre-trained, finetuned, and evaluated the modified GD-MAE with the same setting as \ours. 
As depicted in Table~\ref{tab:two_frames}, directly merging two frames improves the metrics in terms of vehicles and pedestrians, which probably results from the density of target objects being doubled, as shown in Fig.~\ref{fig:veh_example}. 
However, it has a negative impact on cyclists, resulting in a decrease in overall performance. The drop for cyclists might be attributed to the drift of estimated bounding boxes caused by the fast velocity of bicycles and their relatively small dimensions.
In contrast, integrating a historical frame by our method consistently improves overall and class-specific metrics compared to the single-frame baseline.

\begin{table}[t]
\caption{\textbf{Ablation study on model architecture.} The proposed \ourBackbone (e) demonstrates superior performance in terms of overall mAP and mAPH.}
\centering
\scalebox{0.7}{\tablestyle{7pt}{1.0}
\begin{tabular}{lllcccccccc}
\toprule
\multirow{2}{*}{\makecell{Model}} & \multirow{2}{*}{\makecell{Encoder}} & \multirow{2}{*}{\makecell{Fusion}} & \multicolumn{2}{c}{L2 Overall} & \multicolumn{2}{c}{Vehicle} & \multicolumn{2}{c}{Pedestrian} & \multicolumn{2}{c}{Cyclist} \\
\cmidrule(lr){4-5} \cmidrule(lr){6-7} \cmidrule(lr){8-9} \cmidrule(lr){10-11}
         &            &             & mAP & mAPH & AP & APH & AP & APH & AP & APH  \\ \midrule
(a)      & Asymmetric & WCA         & \nd 47.26 & \rd 44.78 & \rd 54.25 & \rd 53.75 & \rd 56.66 & \rd 50.98 & \nd 30.88 & \nd 29.60  \\
(b)      & SimSiam    & WCA         & 45.58 & 42.05 & 53.37 & 52.84 & 53.80 & 44.78 & \rd 29.58 & \rd 28.53  \\
(c)      & Siamese    & WCA$+$WSA   & \rd 47.01 & \nd 45.11 & \nd 55.81 & \nd 55.33 & \fs 58.95 & \fs 54.69 & 26.26 & 25.31 \\
(d)      & Siamese    & WSA         & 43.82 & 40.90 & 53.05 & 52.54 & 54.08 & 48.18 & 24.33 & 21.97 \\
(e) Ours & Siamese    & WCA         & \fs 49.45 & \fs 46.78 & \fs 56.56 & \fs 56.02 & \nd 57.96 & \nd 51.94 & \fs 33.82 & \fs 32.37  \\ 
\bottomrule
\end{tabular}
}
\label{tab:encoder_decoder}
\end{table}

\begin{figure}[t] 
    \centering
    \includegraphics[width=0.1942\linewidth]{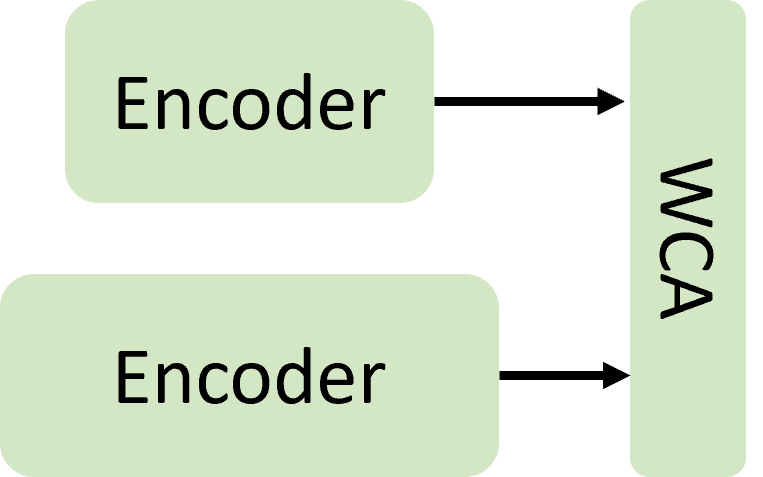}  \hfill
    \includegraphics[width=0.1935\linewidth]{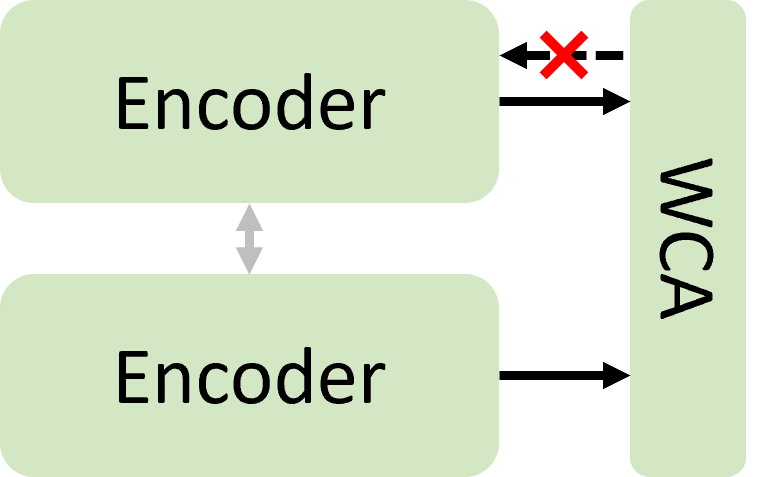} \hfill
    \includegraphics[width=0.2223\linewidth]{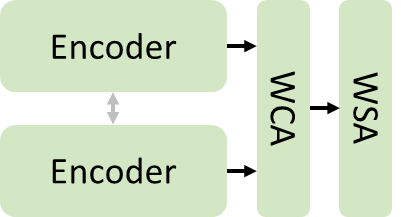} \hfill
    \includegraphics[width=0.2917\linewidth]{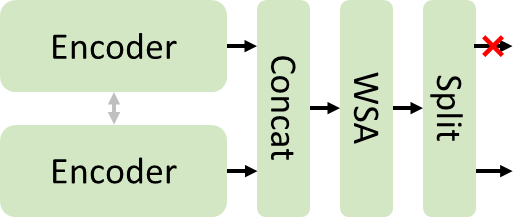}
    \\
    \makebox[0.1942\linewidth]{\scriptsize (a) Asym. Enc.} \hfill
    \makebox[0.1935\linewidth]{\scriptsize (b) SimSiam Enc.} \hfill
    \makebox[0.2223\linewidth]{\scriptsize (c) WCA + WSA} \hfill
    \makebox[0.2917\linewidth]{\scriptsize (d) WSA}
    \\
    \caption{\textbf{Four architecture variants of our SiamWCA backbone.} (a) Asymmetric encoders: the encoder for the previous frame is scaled down. (b) SimSiam-style~\cite{xChen_2021_SimSiam} encoder: one encoder receives no gradient updates. (c) An additional windows-based self-attention is attached as the classic Transformer~\cite{aVaswani_2017_Transformer}. (d) The fusion is implemented by a concat-WSA-split operation.}
    \label{fig:arch_alternatives}
\end{figure}
\boldparagraph{Backbone design.}
The proposed architecture consists of a Siamese encoder and a WCA module.
We ablate these components with alternatives.
\textbf{(a)} Asymmetric encoder: it derives from a Siamese encoder but the encoder for $\mathcal{P}^1$ is scaled down. Consequently, the two encoders no longer share weights.
\textbf{(b)} We employ the SimSiam~\cite{xChen_2021_SimSiam} approach, where one of the two encoders is detached from the computational graph, preventing it from being updated by gradient propagation. The encoder for the current frame is updated by gradients and then shares its weights with the detached encoder.
\textbf{(c)} The design of the Siamese encoder stays unaltered, while a typical cross-self attention module is adopted in a window-based manner. 
\textbf{(d)} The WCA module in our \ourBackbone is replaced with a windowed self-attention (WSA) module. 
However, since self-attention only needs one input, the tokens of both frames have to be concatenated. 
Then, the enhanced tokens of the current frame are split from the output of WSA module.
These variants are depicted in Fig.~\ref{fig:arch_alternatives}.
The superiority of the proposed \ourBackbone is shown in Tab.~\ref{tab:encoder_decoder}.
Variant (c) outperforms \ourBackbone in terms of pedestrians but at the expense of both cyclists' performance and extra parameters.
Therefore, based on the comprehensive comparison, a Siamese encoder and a WCA fusion module are selected.

\boldparagraph{Compatibility.}
To verify that the proposed SSL method \ours is independent of a specific encoder or detector, we replace the encoder SPT~\cite{yang_gd-mae_2023} with two other encoders, \ie SST~\cite{lFan_2022_SST} and SpCNN~\cite{SparseConvNet}, and the detector with a two-stage detector, Graph R-CNN~\cite{hYang_2022_GraphR-CNN}.
Table~\ref{tab:backbone} shows that  T-MAE constantly improves performance by a significant margin, showcasing its compatibility.
\begin{table*}[t]
\caption{\textbf{Ablation study on the encoder and the detector.} The original encoder SPT~\cite{yang_gd-mae_2023} is replaced with SST~\cite{lFan_2022_SST} and SpCNN~\cite{SparseConvNet} and the original detector enterPoint~\cite{tYin_2021_CenterPoint} is replaced with Graph R-CNN~\cite{hYang_2022_GraphR-CNN}. Due to space limits, only overall performance is presented.
}
\centering
\scalebox{0.9}{\tablestyle{6pt}{1.05}
\begin{tabular}{c|ll|cc|cc}
\toprule
\multirow{2}{*}{\makecell{Data\\Amount}} & \multirow{2}{*}{Encoder} & \multirow{2}{*}{Detector} & \multicolumn{2}{c|}{Random Init.} & \multicolumn{2}{c}{\ours (Ours)} \\ 
 \cmidrule(lr){4-5}  \cmidrule(lr){6-7} 
 & & & mAP & mAPH & mAP & mAPH \\ \midrule
\multirow{4}{*}{5\%} 
    & SPT & CenterPoint & 43.68      & 40.29  & \textbf{51.47}  & \textbf{49.46} \\ 
    & SST & CenterPoint & 44.26      & 41.19  & \textbf{51.59}  & \textbf{49.24} \\ 
    & SpCNN & CenterPoint & 46.02    & 43.31  & \textbf{52.08}  & \textbf{49.95} \\ 
    & SPT & Graph R-CNN & 51.18      & 47.27  & \textbf{56.92}  & \textbf{54.70} \\ 
\midrule
\multirow{2}{*}{100\%} 
    & SPT & CenterPoint & 71.30      & 69.13  & \textbf{72.30}  & \textbf{70.52} \\     
    & SPT & Graph R-CNN & 72.76      & 70.04  & \textbf{75.16}  & \textbf{73.50} \\     
\bottomrule
\end{tabular}
}
\label{tab:backbone}
\end{table*}

\boldparagraph{Temporal interval for inference.}
The temporal interval between two frames should be constrained, as previously discussed in Sec.~\ref{sec:pipeline}.
To investigate this matter, we conducted experiments on the Waymo Dataset~\cite{pSun_2020_Waymo},
suggesting that a fixed interval of 0.3 seconds is better than using two consecutive frames. 
\if\RefAppendix1
Additional information is available in Appendix Sec.~\ref{sec:exp_temporal_gap}.
\else
Additional information is available in Appendix Sec. S3.
\fi

\section{Conclusion}
\label{sec:conclusion}

We introduced Temporal Masked Autoencoders (\ours), a novel self-supervised paradigm for LiDAR point cloud pre-training. 
Building upon the single-frame MAE baseline, we incorporated historical frames into the representation using \ourBackbone, with the proposed \ourFusion module playing a pivotal role.
This pre-training enabled the model to acquire robust representations and the ability to capture motion even with very limited labeled data.
By constraining the temporal interval of two frames, we achieved additional performance improvement.
Our experiments on the Waymo dataset and the ONCE dataset demonstrate the effectiveness of our approach by showing improvements over state-of-the-art methods.

\section*{Acknowledgements}
\label{sec:acknowledgement}
This work was financially supported by TomTom, the University of Amsterdam and the allowance of Top consortia for Knowledge and Innovation (TKIs) from the Netherlands Ministry of Economic Affairs and Climate Policy.
This work used the Dutch national e-infrastructure with the support of the SURF Cooperative using grant no. EINF-7940.
Fatemeh Karimi Nejadasl was financed by the University of Amsterdam Data Science Centre.

\bibliographystyle{splncs04}
\bibliography{main}

\if\RefAppendix1
\title{{\ours}: Temporal Masked Autoencoders for Point Cloud Representation Learning}
\def\SuppMatStandalone{0}
\clearpage
{\Large\centering \textbf{Supplementary Material}} 
\beginsupplement

\begin{abstract}
    This supplementary provides more details and analysis of our method.
    The implementation details are illustrated in Sec.~\ref{sec:impl_details}.
    Additional ablation results are provided in Sec.~\ref{sec:additional_ablation}.
    Section~\ref{sec:exp_temporal_gap} discusses the selection of temporal gap during inference.
    More quantitative comparisons with other methods on the Waymo dataset~\cite{pSun_2020_Waymo} are depicted in Sec.~\ref{sec:additional_waymo}.
    The transferability of \ours is verified in Sec.~\ref{sec:transfer} and a comparison with multi-frame non-SSL methods is provided in Sec.~\ref{sec:supp_multiframe}.
    We analyze what the WCA module learns from the \ours pre-training in Sec.~\ref{sec:attention}. 
    Sec.~\ref{sec:training_iterations} provides details about the comparison of finetuning iterations.
    Eventually, more qualitative results and limitations are presented in Sec.~\ref{sec:additional_qua} and Sec.~\ref{sec:limitations}.
\end{abstract}

\section{Implementation Details} \label{sec:impl_details}
We follow the training settings of GD-MAE~\cite{yang_gd-mae_2023}.
Some important configurations are listed in Tab.~\ref{tab:training_details} and Tab.~\ref{tab:dataset_details}.
We use the same masking ratio, the per-pillar number of predicted points $K^O$ and target reconstructed points $K^{GT}$ as GD-MAE because these parameters have negligible effects according to the ablation studies in GD-MAE~\cite{yang_gd-mae_2023}.

\begin{table}[]
\caption{\textbf{Training details.}}
\centering
\scalebox{0.95}{\tablestyle{6pt}{1.0}
\begin{tabular}{l|cc}
\toprule
Config & Pre-training & Finetuning \\
\midrule
optimizer & \multicolumn{2}{c}{AdamW~\cite{AdamW}} \\
optimizer momentum & \multicolumn{2}{c}{$\beta_{1}, \beta_{2} = 0.9, 0.99$} \\
weight decay & \multicolumn{2}{c}{0.01} \\
max learning rate & \multicolumn{2}{c}{0.003} \\
\multirow{2}{*}{\makecell{learning rate scheduler}} & \multicolumn{2}{c}{a cyclic learning rate} \\
                                                    & \multicolumn{2}{c}{with cosine annealing} \\
batch size (Waymo~\cite{pSun_2020_Waymo}) & 4 & 3 \\
batch size (ONCE~\cite{jMao_2021_ONCE}) & 8 & 6 \\
epoch & 12 & 30 \\
masking ratio & 0.75 & - \\
\# predicted points $K^{O}$ & 16 & - \\
\# target reconstructed points $K^{GT}$ & 64 & - \\
\bottomrule
\end{tabular}
}
\label{tab:training_details}
\end{table}

\begin{table}[]
\caption{\textbf{Dataset-specific details.} Finetuning time indicates the duration of finetuning using the entire training set.}
\centering
\scalebox{0.95}{\tablestyle{6pt}{1.0}
\begin{tabular}{l|cc}
\toprule
Config & Waymo~\cite{pSun_2020_Waymo} & ONCE~\cite{jMao_2021_ONCE} \\
\midrule
window size & \multicolumn{2}{c}{ $(8, 8, 1)$ } \\
detection range - x-axis (m) & \multicolumn{2}{c}{$(-74.88, 74.88)$} \\
detection range - y-axis (m) & \multicolumn{2}{c}{$(-74.88, 74.88)$} \\
detection range - z-axis (m) & $(-2, 4)$ & $(-3, 5)$ \\
pillar size (m) & $(0.32, 0.32, 6)$ & $(0.32, 0.32, 8)$ \\
temporal batch & 6 & 3 \\
GPUs & 8 $\times$ Tesla V100 & 4 $\times$ A100 (40GB) \\
Pre-training time (GPU day) & 8 $\times$ 4 & 4 $\times$ 4.5 \\
Finetuning time (GPU day) & 8 $\times$ 5 & 4 $\times$ 0.25 \\
\bottomrule
\end{tabular}
}
\label{tab:dataset_details}
\end{table}

\begin{table}[t]
\caption{\textbf{Ablation experiments on the Waymo dataset~\cite{pSun_2020_Waymo}.}}
\centering
\scalebox{0.9}{\tablestyle{6pt}{1.0}
\begin{tabular}{l|c|cc}
\toprule
\multirow{2}{*}{\makecell{Ablation Target}} & \multirow{2}{*}{\makecell{Setting}} & \multicolumn{2}{c}{L2 Overall} \\
\cmidrule(lr){3-4} 
         & & mAP & mAPH \\
\midrule
\multirow{2}{*}{Two-frame alignment} & baseline        & 41.42 & 37.56 \\
                                     & w/ alignment    & 44.05\up{2.63} & 41.28\up{3.72} \\
\midrule
Data augmentation                    & w/ copy-n-paste~\cite{yYan_2018_SECOND} & 46.76\up{5.34} & 43.93\up{6.37} \\
\midrule
\multirow{3}{*}{Length of temporal batch} & 6$\rightarrow$3 &  45.88\up{4.46} &  41.96\up{4.40} \\
                                          & 6 &  46.76\up{5.34} &  43.93\up{6.37} \\
                                          & 6$\rightarrow$12 &  44.96\up{3.54} &  41.92\up{4.36} \\
\bottomrule
\end{tabular}
}
\label{tab:init_attempts}
\end{table}

\section{Additional Ablation Study} \label{sec:additional_ablation}
Additional results on the Waymo dataset~\cite{pSun_2020_Waymo} are provided to justify design choices and hyperparameter values.
Note that, in this section, we use 20\% of the training set to pre-train our model and use 5\% of the training set to fine-tune it for cost-effective experiments.

\boldparagraph{Two-frame Alignment.}
We start with using \ourBackbone as a baseline and investigate whether it is necessary to align the previous frame to the coordinate system of the current frame.
As shown in Tab.~\ref{tab:init_attempts}, aligning two frames significantly improves the metrics.

\boldparagraph{Data Augmentation.}
Based on the baseline with two-frame alignment, we further investigate the effect of a data augmentation technique, namely copy-n-paste~\cite{yYan_2018_SECOND}.
Specifically, it first generates a ground truth instance database during dataset pre-processing.
Then, during finetuning, several ground truth instances are randomly placed into the scene.
This augmentation boosts the detection rates.
Note that only ground truths in the 5\% split are included in the database, which avoids ground truth leakage.

\boldparagraph{Length of Temporal Batch.}
This hyperparameter is set as 6 for the previous experiments in Tab.~\ref{tab:init_attempts}.
In this ablation, we pre-train and fine-tune the model with different lengths of the temporal batch.
As shown in Tab.~\ref{tab:init_attempts}, setting the length of the temporal batch as 6 achieves the optimal performance.

\section{Temporal interval for inference.} \label{sec:exp_temporal_gap}
\begin{figure}[tb]
  \centering  
  \includegraphics[width=0.98\linewidth]{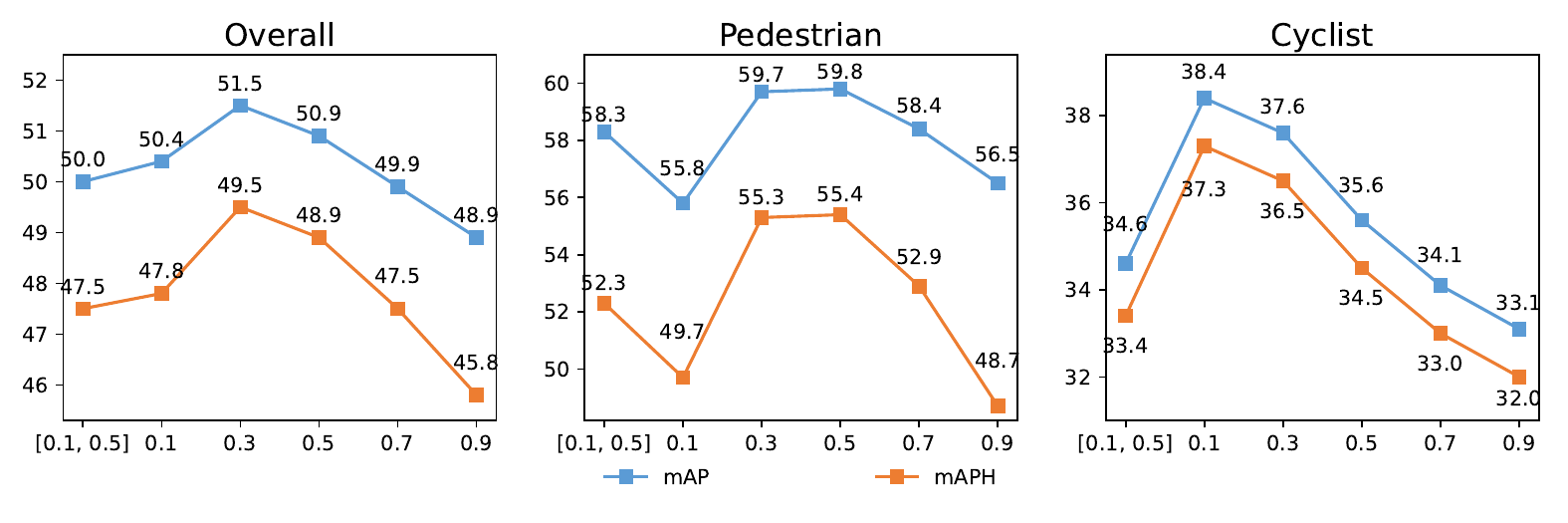}\\[-7pt]
  \caption{\textbf{Temporal interval for inference.} Different temporal intervals were tested for inference. $[0.1, 0.5]$ indicates a random interval between $0.1$ and $0.5$ seconds. A fixed interval ranging from 0.1 to 0.9 seconds was also tested. Overall, a fixed interval of 0.3 seconds works best in our experiments.}
  \label{fig:scatter_temporal_interval}
\end{figure}
This ablation study explores the impact of different temporal intervals on the model performance during inference.
We establish a baseline with a random interval ranging from 0.1 to 0.5 seconds.
In other words, $\mathcal{P}^{t_1}$ is selected randomly from the last five frame of $\mathcal{P}^{t_2}$.
As a comparison, the temporal interval is fixed to values ranging from $0.1$ to $0.9$ seconds.
If there is no satisfactory frame available, the earliest available frame is selected as $\mathcal{P}^{t_1}$.
As shown in Fig.~\ref{fig:scatter_temporal_interval}, The AP and APH for pedestrians achieve relatively high values when the interval is 0.3 and 0.5 seconds, while those for cyclists hit optimal when the interval is 0.1 seconds.
The different optimal intervals could be attributed to the different speeds of the two categories.
Pedestrians can barely move within 0.1 seconds, leading to historical information less useful.
On the contrary, cyclists move relatively faster and thus the same bicycle of two frames is easy to be recognized as two instances if the temporal gap is large.
Fig.~\ref{fig:scatter_temporal_interval} also illustrates that optimal overall performance is achieved when the interval is 0.3 seconds.
Therefore, the previous third frame, which corresponds to a temporal interval of 0.3 seconds, is selected as $\mathcal{P}^{t_1}$ for the Waymo dataset.

\section{Additional Results on Waymo dataset} \label{sec:additional_waymo}
\begin{table*}[t]
\caption{ \textbf{Performance comparison on the Waymo validation set~\cite{pSun_2020_Waymo}.} All methods are fine-tuned with 20\% uniformly sampled frames, namely sampling one frame per five frames. Random initialization denotes training from scratch. Differences between \ours pre-training and random initialization are highlighted in \red{red}. $^{**}$ indicates results from~\cite{yang_gd-mae_2023}. Other results are from AD-PT~\cite{jYuan_2023_AD-PT} or the survey~\cite{bFei_2023_PCSSL_Survey}. Best results are highlighted as \colorbox{colorFst}{\bf first}, \colorbox{colorSnd}{second}, and \colorbox{colorTrd}{third}.}
\centering
\scalebox{0.68}{\tablestyle{6pt}{1.0}
\begin{tabular}{c|l|ll|cccccc}
\toprule
\multirow{2}{*}{\makecell{Data\\Amount}} & \multirow{2}{*}{Initialization} & \multicolumn{2}{c|}{Overall} & \multicolumn{2}{c}{Vehicle} & \multicolumn{2}{c}{Pedestrian} & \multicolumn{2}{c}{Cyclist} \\ \cmidrule(lr){3-4} \cmidrule(lr){5-6} \cmidrule(lr){7-8} \cmidrule(lr){9-10}
 &  & mAP & mAPH & mAP & mAPH & mAP & mAPH & mAP & mAPH \\ \midrule
\multirow{8}{*}{\makecell{20\% \\Uniform\\Sampling}} & Random & \rd 69.03 & \rd 66.48 & \rd 66.61 & \rd 66.15 & \rd 72.65 & \nd 66.76 & 67.83 & 66.54  \\
 & ProposalContrast~\cite{jYin_2022_ProposalContrast}   & 66.67 & 64.20 & 65.22 & 64.80 & 66.40 & 60.49 & 68.48 & 67.38 \\
 & BEV-MAE~\cite{lin_bev-mae_2022}                      & 66.70 & 64.25 & 64.72 & 64.22 & 66.20 & 60.59 & 69.11 & 67.93 \\
 & Occupancy-MAE~\cite{OccupancyMAE}                    & 65.86 & 63.23 & 64.05 & 63.53 & 65.78 & 59.62 & 67.76 & 66.53 \\
 & MAELi-MAE~\cite{krispel_maeli_2023}                  & 65.60 & 63.00 & 64.22 & 63.70 & 65.93 & 59.79 & 66.66 & 65.52  \\
 & GD-MAE~\cite{yang_gd-mae_2023}$^{**}$                & \nd 70.24 & \nd 67.14 & \nd 67.67 & \nd 67.22 & \nd 73.18 & \rd 65.50 & \nd 69.87 & \nd 68.71 \\
 & AD-PT~\cite{jYuan_2023_AD-PT}                        & 67.17 & 64.65 & 65.33 & 64.83 & 67.16 & 61.20 & \rd 69.39 & \rd 68.25 \\
 & \textbf{\ours (Ours)}                                & \fs 70.92\up{1.89} & \fs 69.11\up{2.63} & \fs 68.07 & \fs 67.61 & \fs 74.38 & \fs 70.56 & \fs 70.32 & \fs 69.15 \\
\bottomrule
\end{tabular}
}
\label{tab:waymo_0.20_uniform}
\end{table*}

\begin{table*}[t]
\caption{\textbf{Performance on the Waymo validation set~\cite{pSun_2020_Waymo}.} Results for other methods are taken from MV-JAR~\cite{xu_mv-jar_2023}. All methods are finetuned with Subset 1.}
\centering
\scalebox{0.68}{\tablestyle{6pt}{1.0}
\begin{tabular}{c|l|ll|cccccc}
\toprule
Finetuning     & \multirow{2}{*}{Initialization} & \multicolumn{2}{c|}{Overall} & \multicolumn{2}{c}{Vehicle} & \multicolumn{2}{c}{Pedestrian} & \multicolumn{2}{c}{Cyclist} \\ \cmidrule(l){3-10} 
split  &  & mAP & mAPH & mAP & mAPH & mAP & mAPH & mAP & mAPH \\ \midrule
\multirow{5}{*}{\makecell{5\%\\S1}} & Random            & 49.59 & \rd 46.15 & \rd 54.42 & \rd 53.87 & 52.59 & \rd 44.17 & 41.76 & 40.41 \\
 & PointContrast~\cite{sXie_2020_PointContrast}         & 48.97 & 44.91 & 52.35 & 51.85 & 52.49 & 41.95 & 42.07 & 40.91 \\
 & ProposalContrast~\cite{jYin_2022_ProposalContrast}   & \rd 49.87 & 45.83 & 52.79 & 52.31 & \rd 53.30 & 43.00 & \rd 43.51 & \rd 42.18 \\
 & MV-JAR~\cite{xu_mv-jar_2023}                         & \nd 52.73 & \nd 48.99 & \fs 56.66 & \fs 56.21 & \nd 57.52 & \nd 47.61 & \nd 44.02 & \nd 43.15 \\
 & \textbf{\ours (Ours)}                                & \fs 53.44\up{3.85} & \fs 51.58\up{5.43} & \nd 56.18 & \nd 55.65 & \fs 57.92 & \fs 54.17 & \fs 46.22 & \fs 44.90 \\
\midrule
\multirow{5}{*}{\makecell{10\%\\S1}} & Random           & \rd 57.44 & \rd 54.48 & \nd 59.63 & \nd 59.10 & \rd 60.38 & \rd 53.25 & \rd 52.31 & \rd 51.09 \\
 & PointContrast~\cite{sXie_2020_PointContrast}         & 55.22 & 51.31 & 55.62 & 55.15 & 59.25 & 49.17 & 50.81 & 49.60 \\
 & ProposalContrast~\cite{jYin_2022_ProposalContrast}   & 55.59 & 51.67 & 55.57 & 55.12 & 60.02 & 49.98 & 51.18 & 49.90 \\
 & MV-JAR~\cite{xu_mv-jar_2023}                         & \nd 58.61 & \nd 55.12 & \rd 58.92 & \rd 58.49 & \nd 63.44 & \nd 54.40 & \nd 53.48 & \fs 52.47 \\
 & \textbf{\ours (Ours)}                                & \fs 59.24\up{1.80} & \fs 57.26\up{2.78} & \fs 59.71 & \fs 59.19 & \fs 64.44 & \fs 60.29 & \fs 53.58 & \nd 52.31 \\
\bottomrule
\end{tabular}
}
\label{tab:subset1}
\end{table*}

\begin{table*}[t]
\caption{\textbf{Performance on the Waymo validation set~\cite{pSun_2020_Waymo}.} Results for other methods are taken from MV-JAR~\cite{xu_mv-jar_2023}. All methods are finetuned with Subset 2.}
\centering
\scalebox{0.68}{\tablestyle{6pt}{1.0}
\begin{tabular}{c|l|ll|cccccc}
\toprule
Finetuning   & \multirow{2}{*}{Initialization} & \multicolumn{2}{c|}{Overall} & \multicolumn{2}{c}{Vehicle} & \multicolumn{2}{c}{Pedestrian} & \multicolumn{2}{c}{Cyclist} \\ \cmidrule(l){3-10} 
split &  & mAP & mAPH & mAP & mAPH & mAP & mAPH & mAP & mAPH \\ \midrule
\multirow{5}{*}{\makecell{5\%\\S2}} & Random            & 44.25  & \rd 41.48 & \rd 53.04  & \rd 52.53  & 55.16  & \rd 49.42  & 24.56  & 22.48 \\
 & PointContrast~\cite{sXie_2020_PointContrast}         & 44.48 & 40.55 & 51.87 & 51.37 & 55.36 & 45.03 &  26.22 & 25.24 \\
 & ProposalContrast~\cite{jYin_2022_ProposalContrast}   & \rd 45.21 & 41.45 & 52.29 & 51.82 & \rd 56.23 & 46.28 & \rd 27.10 & \rd 26.24 \\
 & MV-JAR~\cite{xu_mv-jar_2023}                         & \nd 47.93 & \nd 44.50 & \nd 56.22 & \nd 55.78 & \nd 58.80 & \nd 49.77 & \nd 28.75 & \nd 27.95 \\
 & \textbf{\ours (Ours)}                                & \fs 49.01\up{4.75}  & \fs 46.21\up{4.74}  & \fs 56.50  & \fs 55.95  & \fs 60.70  & \fs 54.70  & \fs 29.82  & \fs 27.98 \\
\midrule
\multirow{5}{*}{\makecell{10\%\\S2}} & Random           & \rd 56.81 & \rd 53.97 & \nd 59.49 & \nd 58.98 & \rd 62.44 & \rd 55.54 & \rd 48.50 & \rd 47.40 \\
 & PointContrast~\cite{sXie_2020_PointContrast}         & 54.80 & 51.02 & 55.41 & 54.95 & 60.56 & 50.86 & 48.44 & 47.24 \\
 & ProposalContrast~\cite{jYin_2022_ProposalContrast}   & 54.77 & 51.09 & 55.64 & 55.20 & 60.54 & 51.16 & 48.14 & 46.92 \\
 & MV-JAR~\cite{xu_mv-jar_2023}                         & \nd 58.29 & \nd 54.99 & \rd 59.17 & \rd 58.74 & \nd 64.58 & \nd 56.02 & \fs 51.12 & \fs 50.20 \\ 
 & \textbf{\ours (Ours)}                                & \fs 58.64\up{1.83} & \fs 56.71\up{2.73} & \fs 60.20 & \fs 59.70 & \fs 66.10 & \fs 61.79 & \nd 49.62 & \nd 48.63 \\
\bottomrule
\end{tabular}
}
\label{tab:subset2}
\end{table*}

\begin{table*}[t]
\caption{\textbf{Average performance on the Waymo validation set~\cite{pSun_2020_Waymo}}, averaged across the models finetuned with Subset 0$\sim$2. Results for other methods are taken from MV-JAR~\cite{xu_mv-jar_2023}. \ours consistently enhances model performance with a notable margin compared to random initialization.
}
\centering
\scalebox{0.68}{\tablestyle{6pt}{1.0}
\begin{tabular}{c|l|ll|cccccc}
\toprule
 Finetuing   & \multirow{2}{*}{Initialization} & \multicolumn{2}{c|}{Overall} & \multicolumn{2}{c}{Vehicle} & \multicolumn{2}{c}{Pedestrian} & \multicolumn{2}{c}{Cyclist} \\ \cmidrule(l){3-10} 
 split &  & mAP & mAPH & mAP & mAPH & mAP & mAPH & mAP & mAPH \\ \midrule
 \multirow{5}{*}{\makecell{5\%\\S0$\sim$S2}} & Random   & 45.84  & 42.64  & \rd 53.84  & \rd 53.30  & 53.73  & \rd 46.12  & 29.95  & 28.50 \\
 & PointContrast~\cite{sXie_2020_PointContrast}         & 46.26 & 42.25 & 52.11 & 51.61 & 53.84 & 43.40 & 32.82 & 31.74 \\
 & ProposalContrast~\cite{jYin_2022_ProposalContrast}   & \rd 47.23 & \rd 43.28 & 52.58 & 52.10 & \rd 54.61 & 44.37 & \rd 34.50 & \rd 33.38 \\
 & MV-JAR~\cite{xu_mv-jar_2023}                         & \nd 50.39 & \nd 46.72 & \nd 56.45 & \nd 56.00 & \nd 57.99 & \nd 48.36 & \nd 36.74 & \nd 35.81 \\
 & \textbf{\ours (Ours)}                                & \fs 51.31\up{5.47} & \fs 49.08\up{6.44} & \fs 56.60 & \fs 56.08 & \fs 59.44 & \fs 54.72 & \fs 37.88 & \fs 36.46 \\
\midrule
\multirow{5}{*}{\makecell{10\%\\S0$\sim$S2}} & Random   & \rd 56.77 & \rd 53.86 & \nd 59.63 & \nd 59.12 & \rd 60.97 & \rd 53.94 & \rd 49.70 & \rd 48.52 \\
 & PointContrast~\cite{sXie_2020_PointContrast}         & 54.57 & 50.75 & 55.26 & 54.80 & 59.85 & 50.05 & 48.61 & 47.41 \\
 & ProposalContrast~\cite{jYin_2022_ProposalContrast}   & 54.75 & 50.96 & 55.47 & 55.01 & 60.19 & 50.51 & 48.60 & 47.37 \\
 & MV-JAR~\cite{xu_mv-jar_2023}                         & \nd 58.12 & \nd 54.72 & \rd 58.84 & \rd 58.41 & \nd 63.77 & \nd 55.03 & \nd 51.74 & \nd 50.73 \\
 & \textbf{\ours (Ours)}                                & \fs 59.27\up{2.50} & \fs 57.32\up{3.46} & \fs 60.06 & \fs 59.55 & \fs 65.26 & \fs 61.06 & \fs 52.50 & \fs 51.34 \\
\bottomrule
\end{tabular}
}
\label{tab:waymo_average}
\end{table*}

We explore two methods to obtain a subset of the training set for finetuning, but all models are evaluated on the same validation set.
\textbf{Uniform Sampling} is commonly used~\cite{jYin_2022_ProposalContrast, yang_gd-mae_2023, hLiang_2021_ExploringGeometryawareContrast}. Specifically, frames from all sequences are concatenated, from which the target frames are sampled with a fixed interval.
\textbf{Data-efficient Benchmark}~\cite{xu_mv-jar_2023} selects a certain number of sequences as a training subset instead of uniform sampling frames, which aims to solve the data diversity issue existing in uniform sampling. More specifically, while finetuned on uniformly sampled frames, models always converge to a similar performance if they are finetuned for adequate iterations no matter how many percentages of labelled data are used. As a result, fewer data do not lead to a shorter finetuning time.
The data-efficient benchmark enables models to converge with much fewer iterations.
\if\SuppMatStandalone1
Therefore, to evaluate pre-trained representation more efficiently, we conduct experiments on the data-efficient benchmark, as presented in Tab.~1.
\else
Therefore, to evaluate pre-trained representation more efficiently, we conduct experiments on the data-efficient benchmark, as presented in Tab.~\ref{tab:waymo_main}.
\fi

\boldparagraph{Results Analysis.}
Under the uniform sampling setting, we finetuned the \ours pre-trained model with 20\% uniformly sampled frames.
As shown in Tab.~\ref{tab:waymo_0.20_uniform}, \ours outperforms other methods, which aligns with the statement made in the main paper.
Under the data-efficient setting, the 5\% and 10\% splits contain limited samples, which may lead to performance variance.
Therefore, 
\if\SuppMatStandalone1
except for the split 0 used for Tab.~1,
\else
except for the split 0 used for Tab.~\ref{tab:waymo_main},
\fi
we also compare models in Tab.~\ref{tab:subset1} and Tab.~\ref{tab:subset2} when they are finetuned with split 1 and split 2, both of which are provided by MV-JAR~\cite{xu_mv-jar_2023} as well.
Furthermore, the average results are presented in Tab.~\ref{tab:waymo_average}.
In conclusion, it can be observed that the model performance exhibits variation when finetuned with different subsets.
However, \ours pre-training consistently improves model performance compared to random initialization.
\ours pre-training surpasses other SSL methods in terms of most class-specific metrics and all overall metrics.
Notably, \ours outperforms other methods in terms of all metrics while the results are averaged (see Tab.~\ref{tab:waymo_average}).

\section{Transferring performance.} \label{sec:transfer}
To assess the transferability of \ours pre-training, we pre-train the model on the ONCE dataset~\cite{jMao_2021_ONCE} and then fine-tune it on the Waymo dataset~\cite{pSun_2020_Waymo}.
Table~\ref{tab:once2waymo} demonstrates that T-MAE offers robust generalizability and transferability.
\begin{table}[b]
\caption{\textbf{Transferability comparison.} }
\centering
\scalebox{0.9}{\tablestyle{6pt}{1.0}
\begin{tabular}{llcccc}
\toprule
Initialization & Data & Overall & Vehicle & Pedestrian & Cyclist \\
\midrule
Random         & -    & \rd 66.48   & \rd 66.15   & \nd 66.76      & \rd 66.54   \\
GD-MAE~\cite{yang_gd-mae_2023}  & ONCE & \nd 66.61   & \nd 67.18   & \rd 64.82      & \nd 67.83   \\
\textbf{\ours (Ours)}          & ONCE & \fs 67.86   & \fs 68.06   & \fs 67.25      & \fs 68.26  \\
\bottomrule
\end{tabular}
}
\label{tab:once2waymo}
\end{table}

\section{Multi-frame comparison with non-SSL methods.} \label{sec:supp_multiframe}
Since there are no existing multi-frame SSL methods, we compare our approach with robust non-SSL baselines that utilize multi-frame as inputs.
Table~\ref{tab:multiframe} demonstrates our method surpasses other supervised methods while requiring only two frames, highlighting the effectiveness of our method.
\begin{table*}[t]
\caption{\textbf{Multi-frame comparison with non-SSL methods.} ``-'' indicates not available.}
\centering
\scalebox{0.9}{\tablestyle{6pt}{1.0}
\scriptsize
\begin{tabular}{lccccccccc}
\toprule
\multirow{2}{*}{Method}  & \multirow{2}{*}{Frames} & \multicolumn{2}{c}{L2 Overall} & \multicolumn{2}{c}{Vehicle} & \multicolumn{2}{c}{Pedestrian} & \multicolumn{2}{c}{Cyclist} \\ 
\cmidrule(lr){3-4} \cmidrule(lr){5-6} \cmidrule(lr){7-8} \cmidrule(lr){9-10}
                                         &      & mAP   & mAPH  & mAP    & mAPH   & mAP    & mAPH   & mAP    & mAPH  \\ 
\midrule
3D-MAN~\cite{zYang_2021_3D-MAN}          & 16   & -     &  -    & 67.6  & 67.1  & 62.6  & 59.0 & -     & -      \\
CenterPoint~\cite{tYin_2021_CenterPoint} & 4    & \nd 70.8  & \nd 69.4  & \nd 69.1  & \nd 68.6  & \rd 71.7  & \rd 68.6  & \nd 71.6  & \fs 70.9  \\
SST~\cite{lFan_2022_SST}                 & 3    & -     &  -    & \rd 68.5  & \rd 68.1  & \nd 75.1  & \nd 70.9 & - & -        \\
\textbf{\ours (Ours)}                    & 2    & \fs 72.3 & \fs 70.5 & \fs 69.4 & \fs 68.9 & \fs 75.8 & \fs 72.0 & \fs 71.8 & \nd 70.7  \\
\bottomrule
\end{tabular}
}
\label{tab:multiframe}
\end{table*}

\begin{figure*}[t] \centering
    \includegraphics[width=0.44\textwidth]{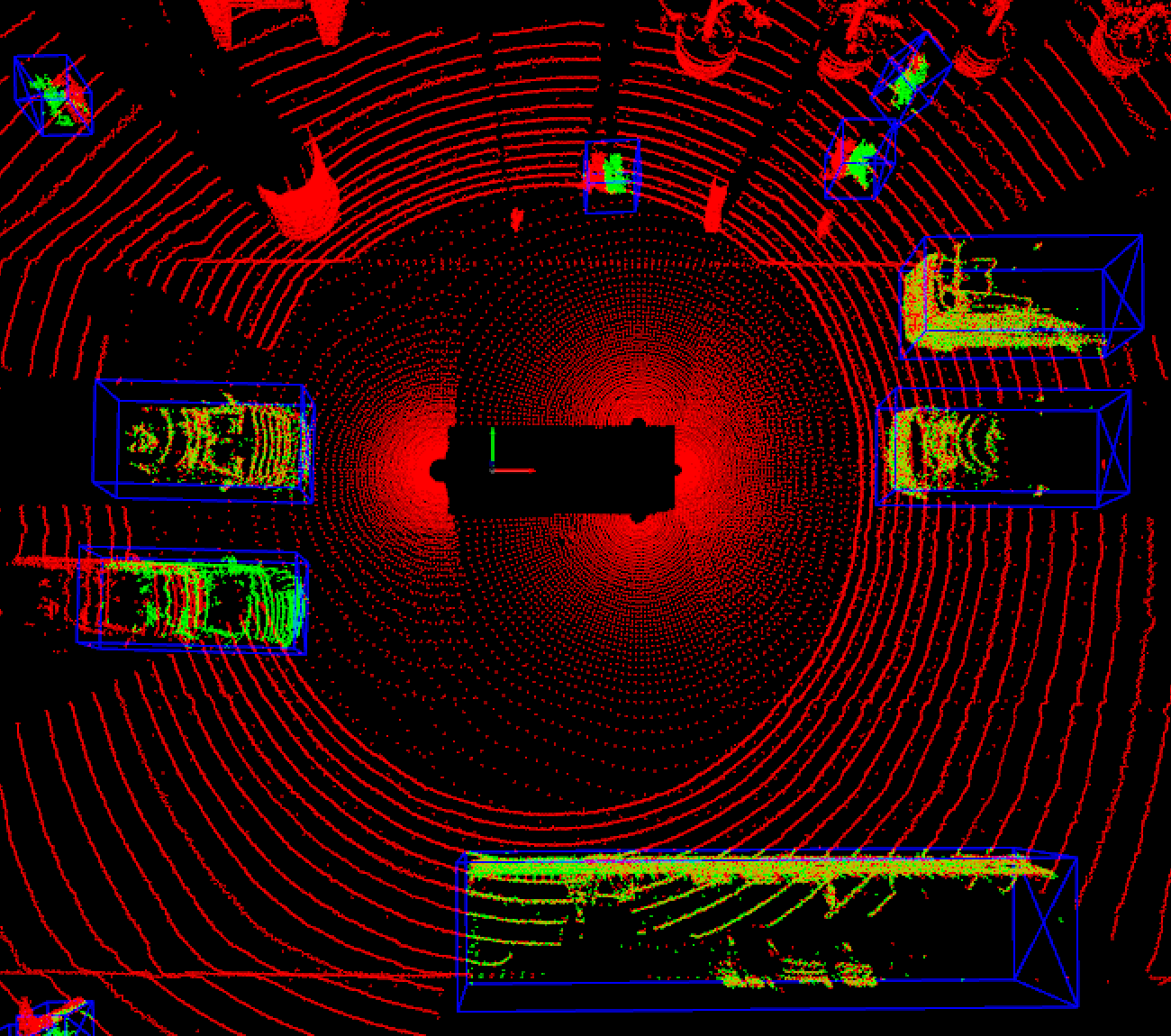} \hfill
    \includegraphics[width=0.44\textwidth]{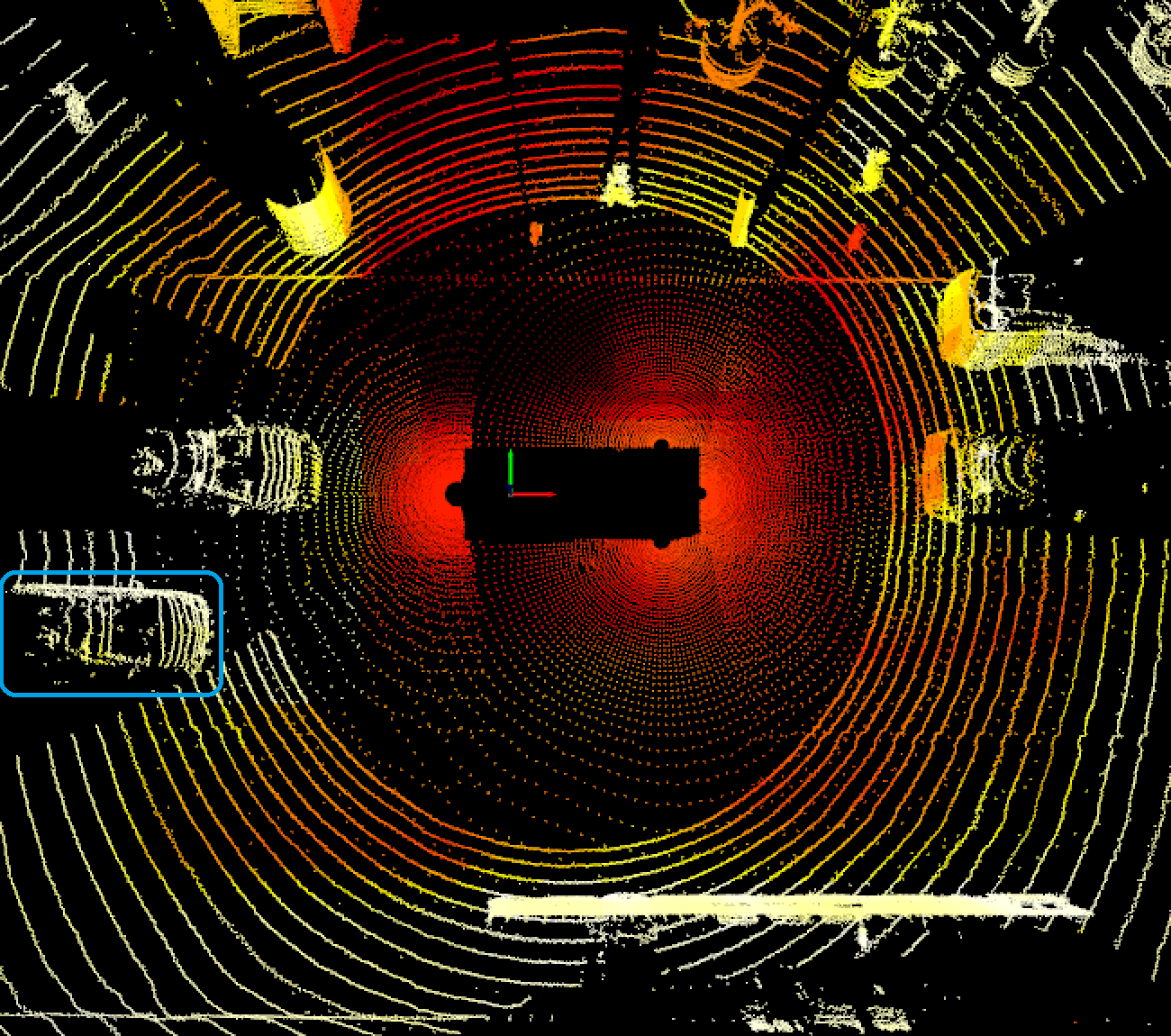} 
    \includegraphics[width=0.064\textwidth]{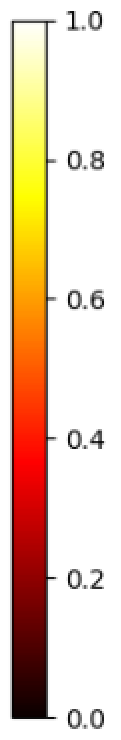}
    \\
    \makebox[0.44\textwidth]{\small (a) Input} \hfill
    \makebox[0.504\textwidth]{\small (b) Attention scores for prior frame} 
    \\ 
    \caption{\textbf{Visualization of attention scores. (a)} The input consists of two point clouds: the entire previous frame (red points) and the current frame (green points) that contains only points within the ground-truth bounding boxes. Note that, the blue bounding boxes in (a) serve solely for visualization purposes and do not function as input. 
    \textbf{(b)} The pillar-wise attention scores are visualized. The attention scores are derived from the \ourFusion module and mapped to a colormap ranging from black to white. The primary attention is placed on the target objects from the previous frame. This implies that the \ourFusion is able to locate corresponding objects. Notably, \ourFusion is capable of accurately trace to the source object, which is manually indicated by a skyblue box, even in cases where the vehicle is moving.}
    \label{fig:attn_map}
\end{figure*}

\section{Attention Learned by T-MAE Pre-training} \label{sec:attention}
In the main paper, \ours pre-trained weights are loaded to both the Siamese encoder and WCA module.
We conduct an ablation study where only the Siamese encoder is initialized by the pre-trained weights and the WCA module is randomly intialized.
As shown in Tab.~\ref{tab:wca_random_init}, initializing WCA with \ours pre-trained weights significantly improves mAPH for pedestrians, which indicates a way better direction detection for pedestrians.
It also boosts metrics for cyclists with a big margin.
\begin{table}[tb]
\caption{\textbf{Ablation on WCA initialization.} SE stands for the Siamese Encoder.}
\centering
\scalebox{0.7}{\tablestyle{6pt}{1.0}
\begin{tabular}{l|cc|ll|cccccc}
\toprule
\multirow{2}{*}{Initialization} & \multicolumn{2}{c|}{Pre-trained} & \multicolumn{2}{c|}{Overall} & \multicolumn{2}{c}{Vehicle} & \multicolumn{2}{c}{Pedestrian} & \multicolumn{2}{c}{Cyclist} \\ 
\cmidrule(lr){2-3} \cmidrule(lr){4-5} \cmidrule(lr){6-7} \cmidrule(lr){8-9} \cmidrule(lr){10-11}
& { }SE & WCA & mAP & mAPH & mAP & mAPH & mAP & mAPH & mAP & mAPH \\ 
\midrule
Random              & \redx & \redx               & \rd 43.68 & \rd 40.29 & \rd 54.05 & \rd 53.50 & \rd 53.45 & \rd 44.76 & \rd 23.54 & \rd 22.61 \\
Partially random    & \greencheck & \redx         & \nd 48.19\up{4.51} & \nd 45.16\up{4.87} & \nd 55.91 & \nd 55.38 & \nd 56.29 & \nd 48.74 & \nd 32.38 & \nd 31.36  \\
T-MAE (ours)        & \greencheck & \greencheck   & \fs 51.47\up{7.79} & \fs 49.46\up{9.17} & \fs 57.13 & \fs 56.63 & \fs 59.69 & \fs 55.28 & \fs 37.61 & \fs 36.48 \\ 

\bottomrule
\end{tabular}
}
\label{tab:wca_random_init}
\end{table}

To further understand the knowledge acquired by the WCA module during pre-training, we employ attention visualization to identify critical regions of the previous frame.
In particular, the entire previous frame and the target objects of the current frame are input into the \ourBackbone that loads \ours pre-trained weights.
More precisely, the input to the model for the current frame solely consists of points contained within the 3D ground-truth bounding boxes.
The purpose of removing other points is to identify the specific areas of emphasis within the \ourFusion module when the queries are solely target objects, \eg vehicles, pedestrians, and cyclists.
As illustrated in Fig.~\ref{fig:attn_map} (b), the attention is mainly attached to target objects, indicating that the \ourFusion successfully detects and localizes these entities in the previous frame.
Furthermore, the presence of vehicular action is perceptible even if the \ourFusion module is only trained with unlabeled data, indicating the effectiveness of our \ours pre-training strategy.

\section{Comparison of training iterations} \label{sec:training_iterations}

\begin{table*}[t]
\caption{\textbf{Quantitative details on comparison of finetuning iterations.} Our scheme requires much less iterations for finetuning compared to MV-JAR~\cite{xu_mv-jar_2023}.}
\centering
\scalebox{0.6}{\tablestyle{6pt}{1.0}
\begin{tabular}{lccccccc}
\toprule
Method & \makecell{Data\\ amount} & Epochs & \makecell{Number of input scans\\($\times 10^3$ per epoch)} & \makecell{Total iterations\\($\times 10^5$)} & \makecell{Number of GT used\\($\times 10^3$ per epoch)} & \makecell{L2 mAPH\\(Overall)} & \makecell{L2 mAPH\\(Pedestrian)} \\  \midrule
\multirow{3}{*}{MV-JAR~\cite{xu_mv-jar_2023}} & 5\%  & 72  & 7.9   & 5.70   & 7.9   & 46.68 & 47.69 \\
                                              & 10\% & 60  & 15.8  & 9.50   & 15.8  & 54.06 & 54.66 \\
                                              & 20\% & 48  & 31.6  & 15.19  & 31.6  & 59.15 & 59.02 \\ \midrule
\multirow{3}{*}{Random/T-MAE}                 & 5\%  & 30  & 7.9   & 2.37   & 4.0   & 40.29/49.46 & 44.76/55.28 \\
                                              & 10\% & 30  & 15.8   & 4.75  & 7.9   & 53.13/57.99 & 53.04/61.10 \\
                                              & 20\% & 30  & 31.6  & 9.50   & 15.8  & 57.61/61.80 & 58.41/64.66 \\ \bottomrule
\end{tabular}
}
\label{tab:details_fig1}
\end{table*}

MV-JAR~\cite{xu_mv-jar_2023} applies varying numbers of epochs during finetuning, depending on the data amount.
Rather than varying numbers, we employ a predetermined number of epochs, leading to significantly fewer iterations for finetuning without a performance drop.
Detailed information is presented in Tab.~\ref{tab:details_fig1}.
The \emph{total number of iterations} is calculated by multiplying \emph{the number of input scans} by \emph{the number of epochs}.
\if\SuppMatStandalone1
As depicted in Fig.~1 and Tab.~\ref{tab:details_fig1}, \ours outperforms MV-JAR~\cite{xu_mv-jar_2023} while requiring a smaller number of finetuning iterations.
\else
As depicted in Fig.~\ref{fig:fig1} and Tab.~\ref{tab:details_fig1}, \ours outperforms MV-JAR~\cite{xu_mv-jar_2023} while requiring a smaller number of finetuning iterations.
\fi

\section{Qualitative Results} \label{sec:additional_qua}
Fig.~\ref{fig:supp_waymo_qua} shows the qualitative results of our method on the Waymo dataset~\cite{pSun_2020_Waymo}.
Fig.~\ref{fig:supp_once_qua1} and Fig.~\ref{fig:supp_once_qua2} illustrates qualitative results on the ONCE datset~\cite{jMao_2021_ONCE}.
Our method generates accurate bounding boxes even if the scene is complex.

\section{Limitations} \label{sec:limitations}
While our work has achieved encouraging results, there is space for further improvements.
For instance, the alignment of two frames relies on their transformation matrices; without proper alignment, there is a drop in performance.
If this could be eliminated, the proposal would be more efficient.
Another shortcoming is that the transformer blocks are computationally heavy, which restricts the window size.
This leads to a smaller receptive field and thus makes the network easy to lose track of fast-moving objects.
The transformer-based encoder also increases the inference time, which was tested on a single NVIDIA RTX 3090 and measured as 429 ms per frame. Consequently, the network is not real-time.

\begin{figure*}[] \centering
    \includegraphics[width=0.95\linewidth]{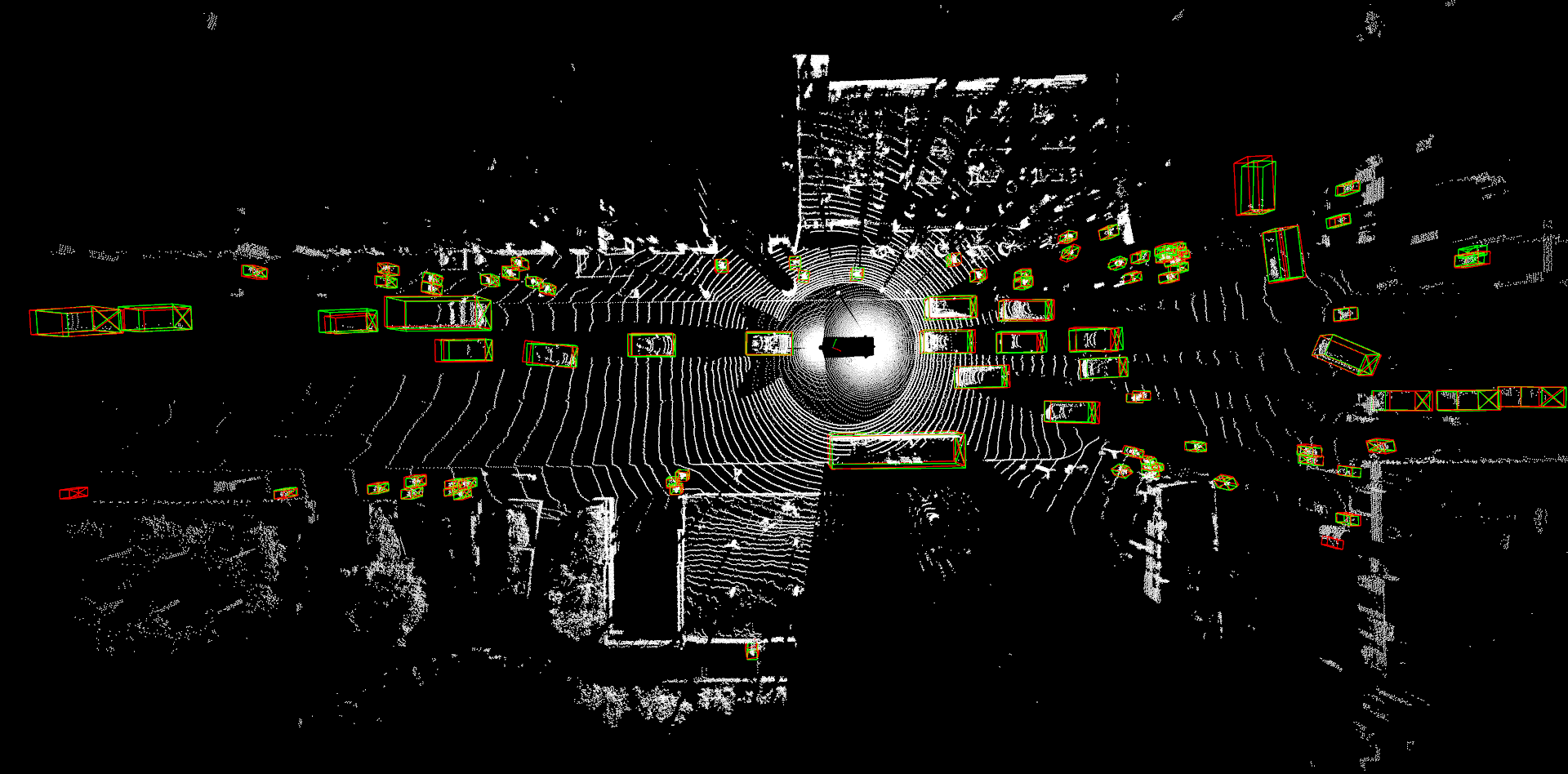} \\
    \includegraphics[width=0.95\linewidth]{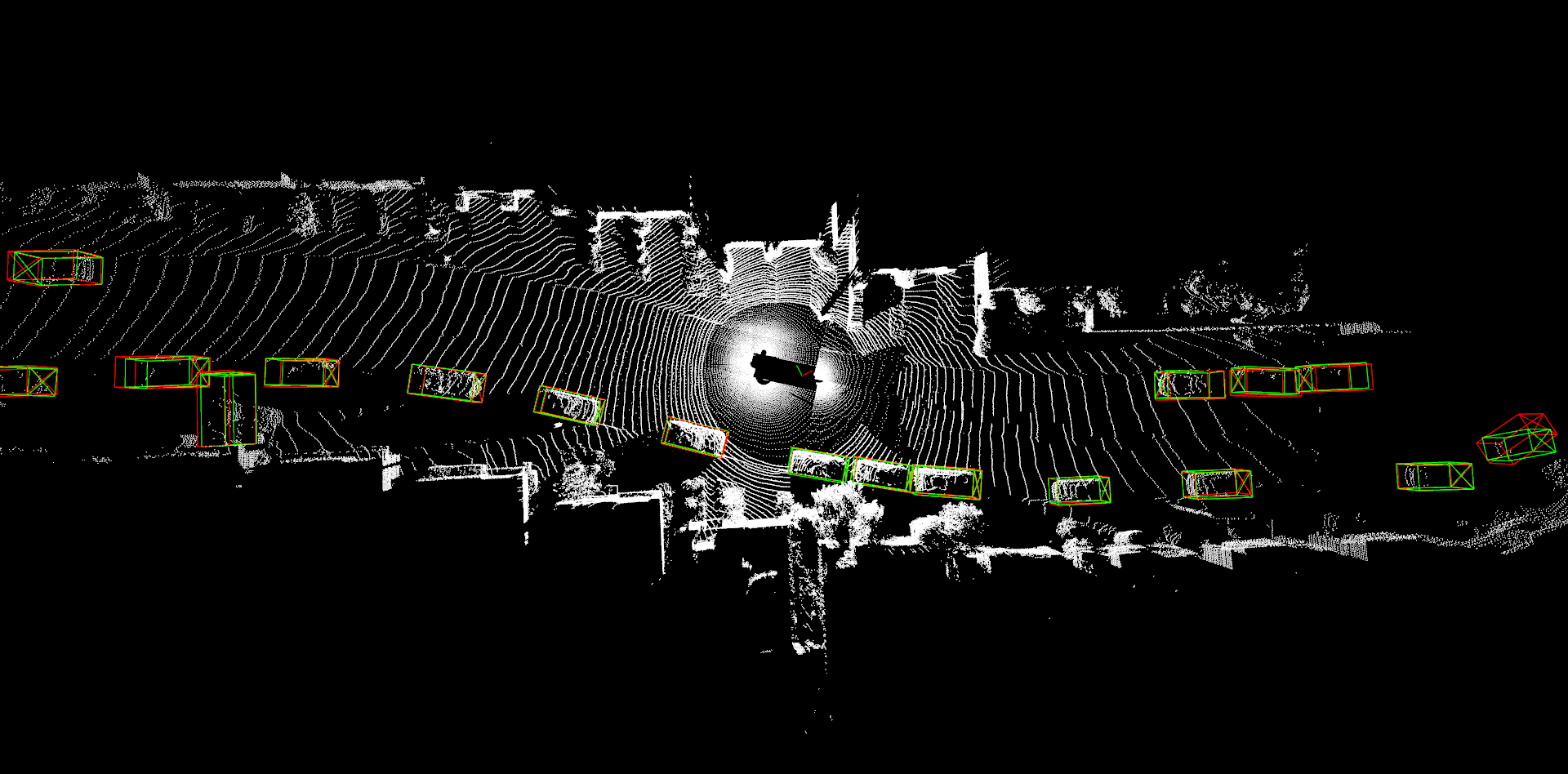} \\
    \includegraphics[width=0.95\linewidth]{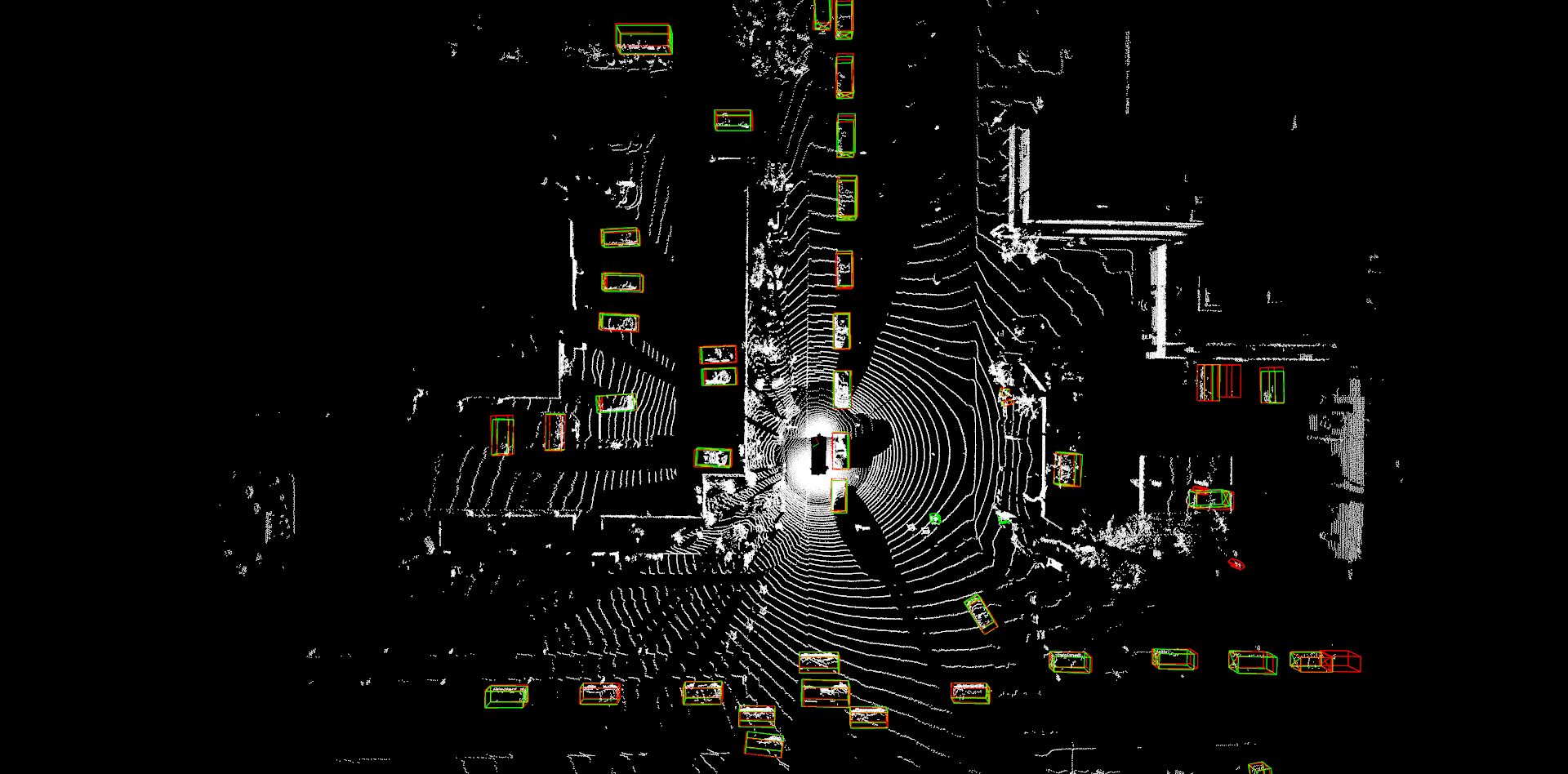}
    \caption{\textbf{Qualitative results on the Waymo dataset~\cite{pSun_2020_Waymo}.} We depict ground truth and predictions as boxes colored in red and green for several exemplary scenes.} 
    \label{fig:supp_waymo_qua}
\end{figure*}
\begin{figure*}[] \centering
    \includegraphics[width=0.95\linewidth]{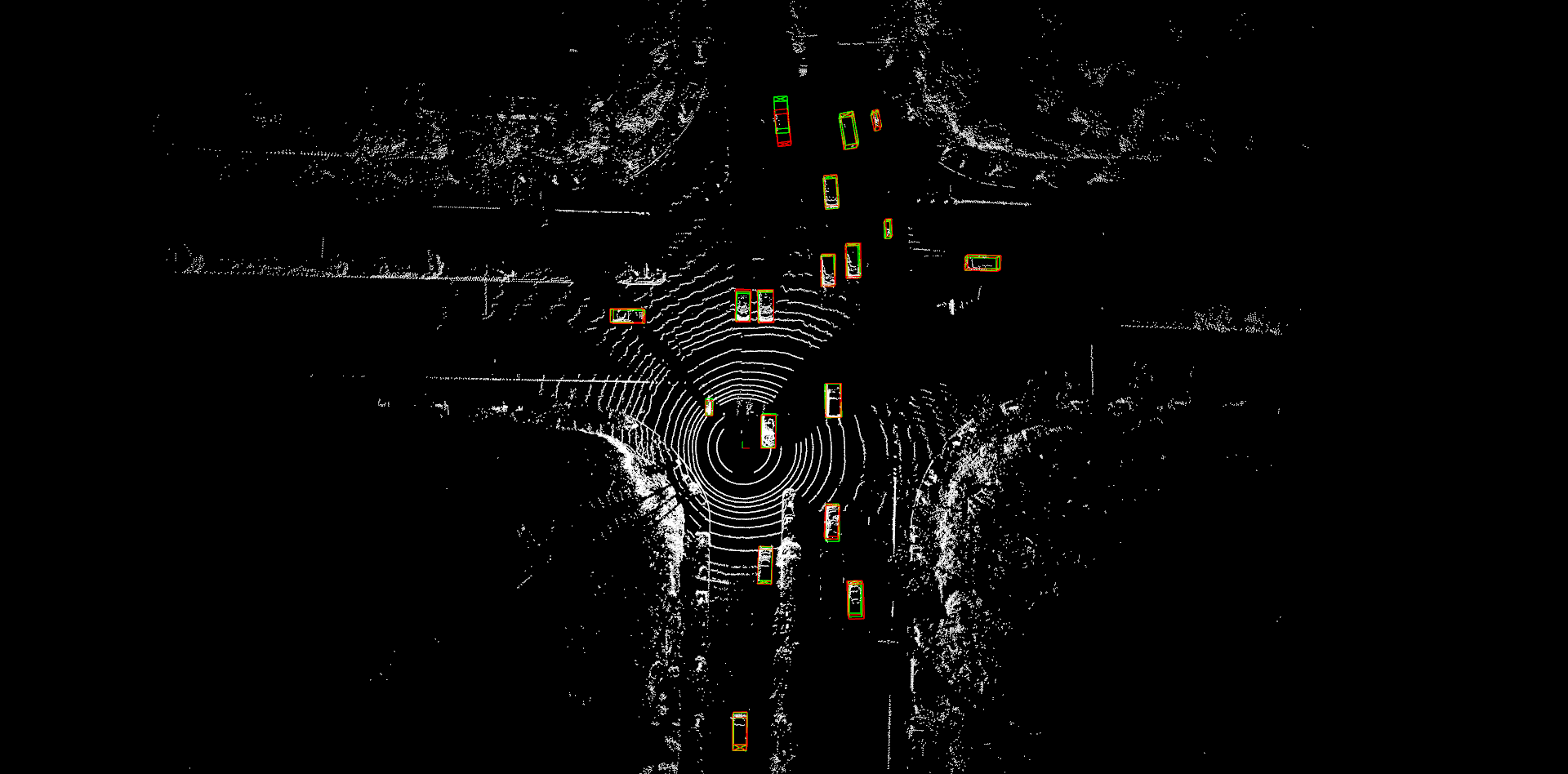} \\
    \includegraphics[width=0.95\linewidth]{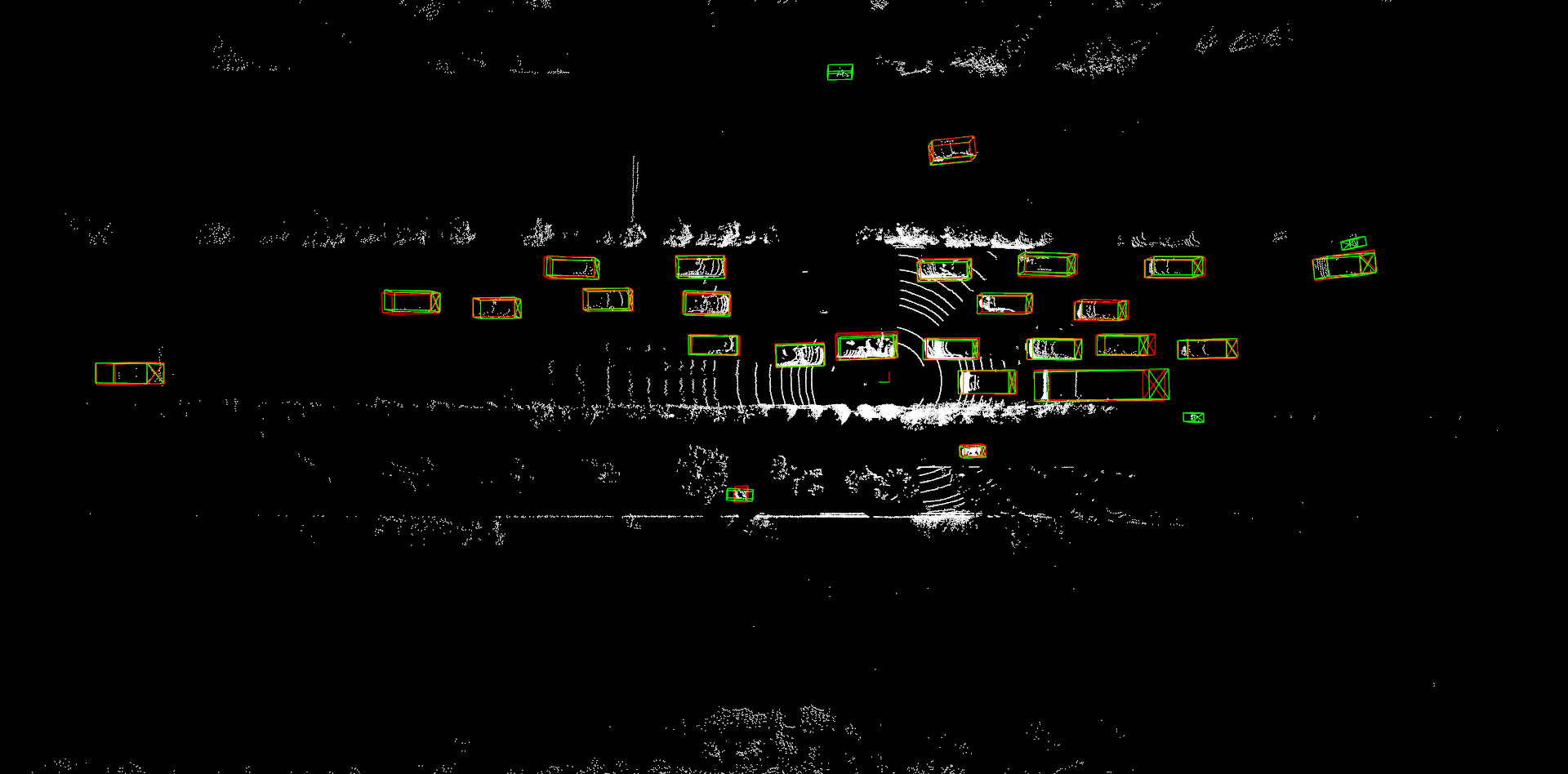} \\
    \includegraphics[width=0.95\linewidth]{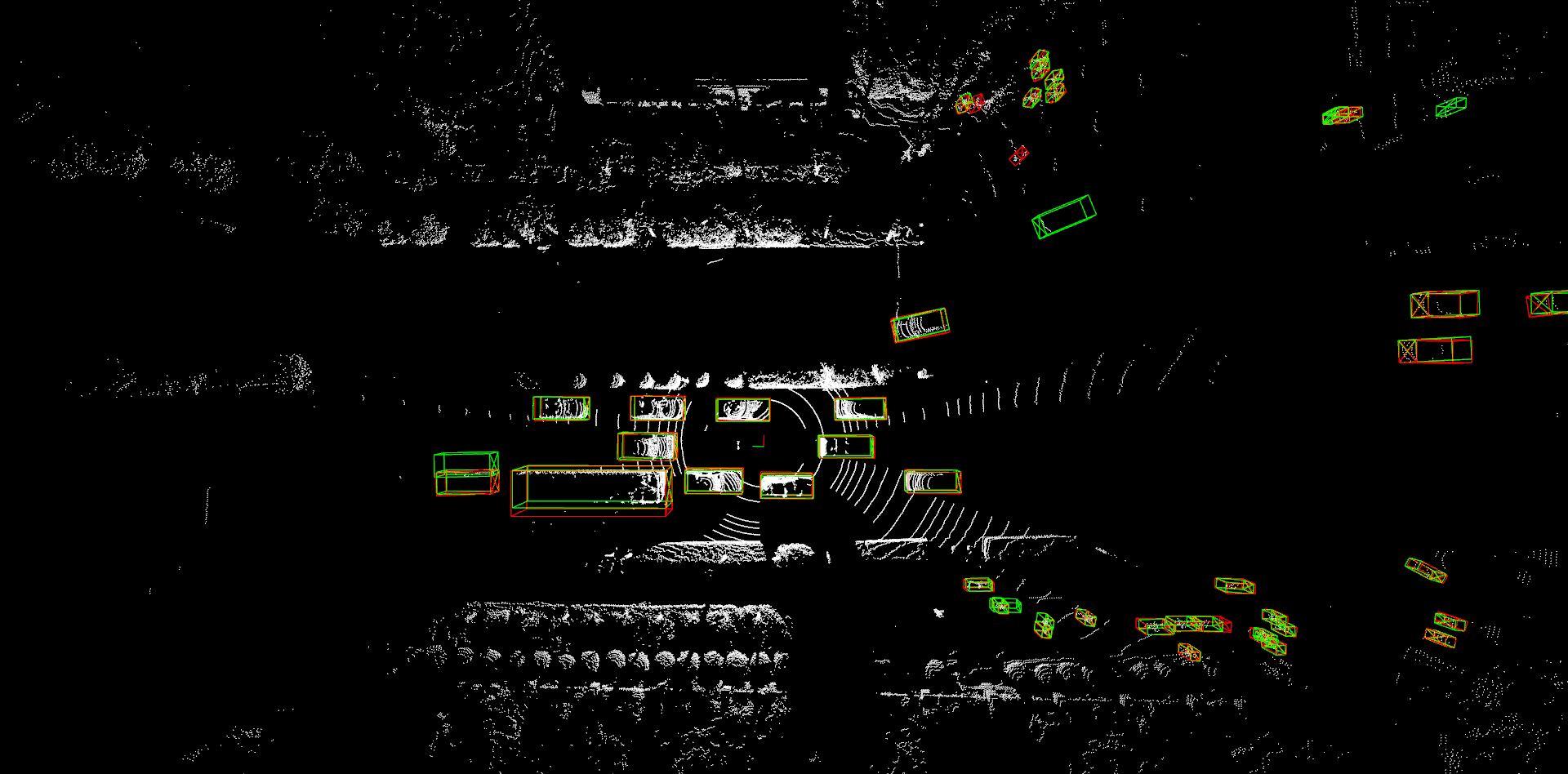}
    \caption{\textbf{Qualitative results on the ONCE  dataset~\cite{jMao_2021_ONCE}.} We depict ground truth and predictions as boxes colored in red and green for several exemplary scenes.} 
    \label{fig:supp_once_qua1}
\end{figure*}

\begin{figure*}[] \centering
    \includegraphics[width=0.95\linewidth]{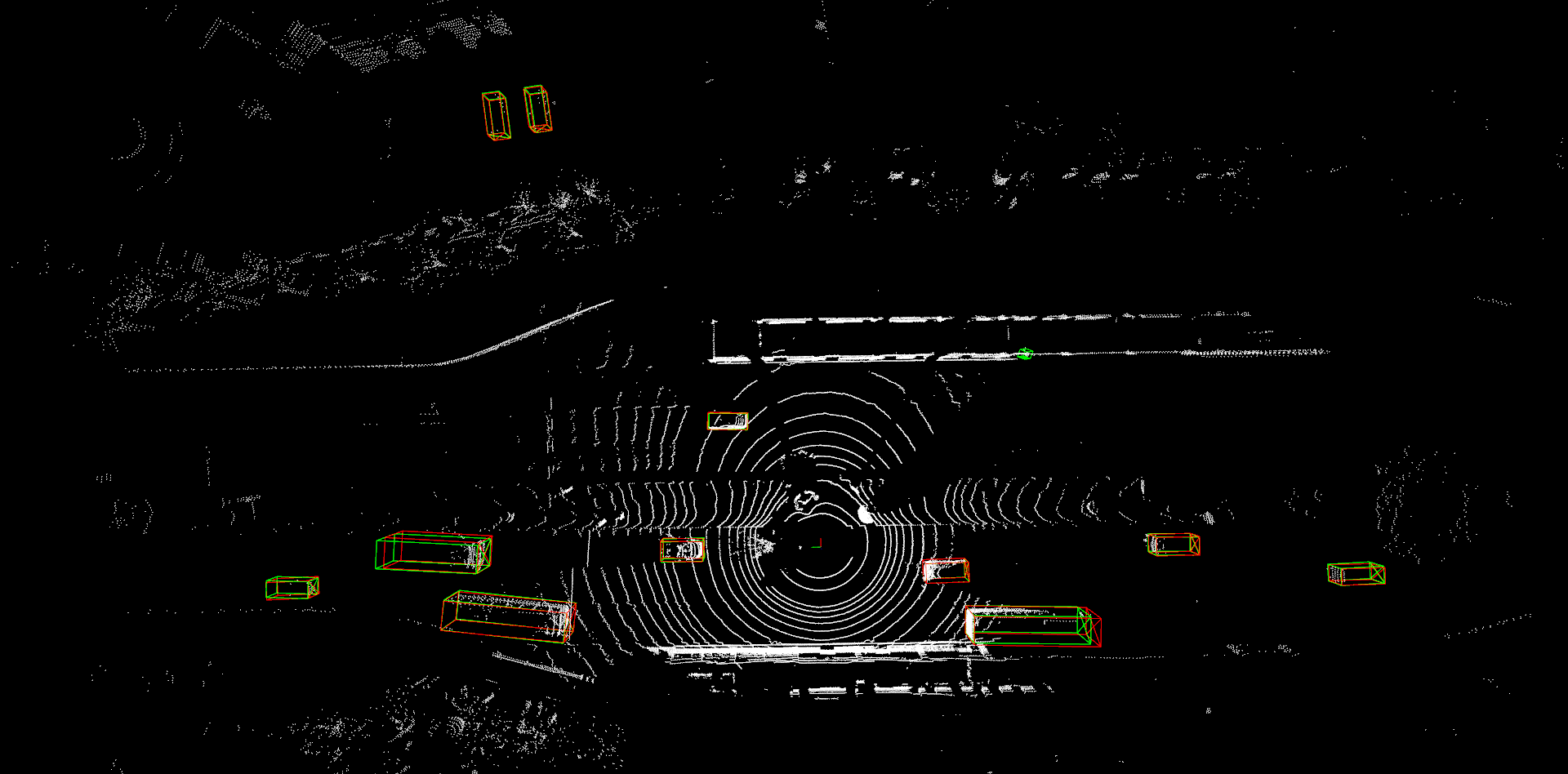} \\
    \includegraphics[width=0.95\linewidth]{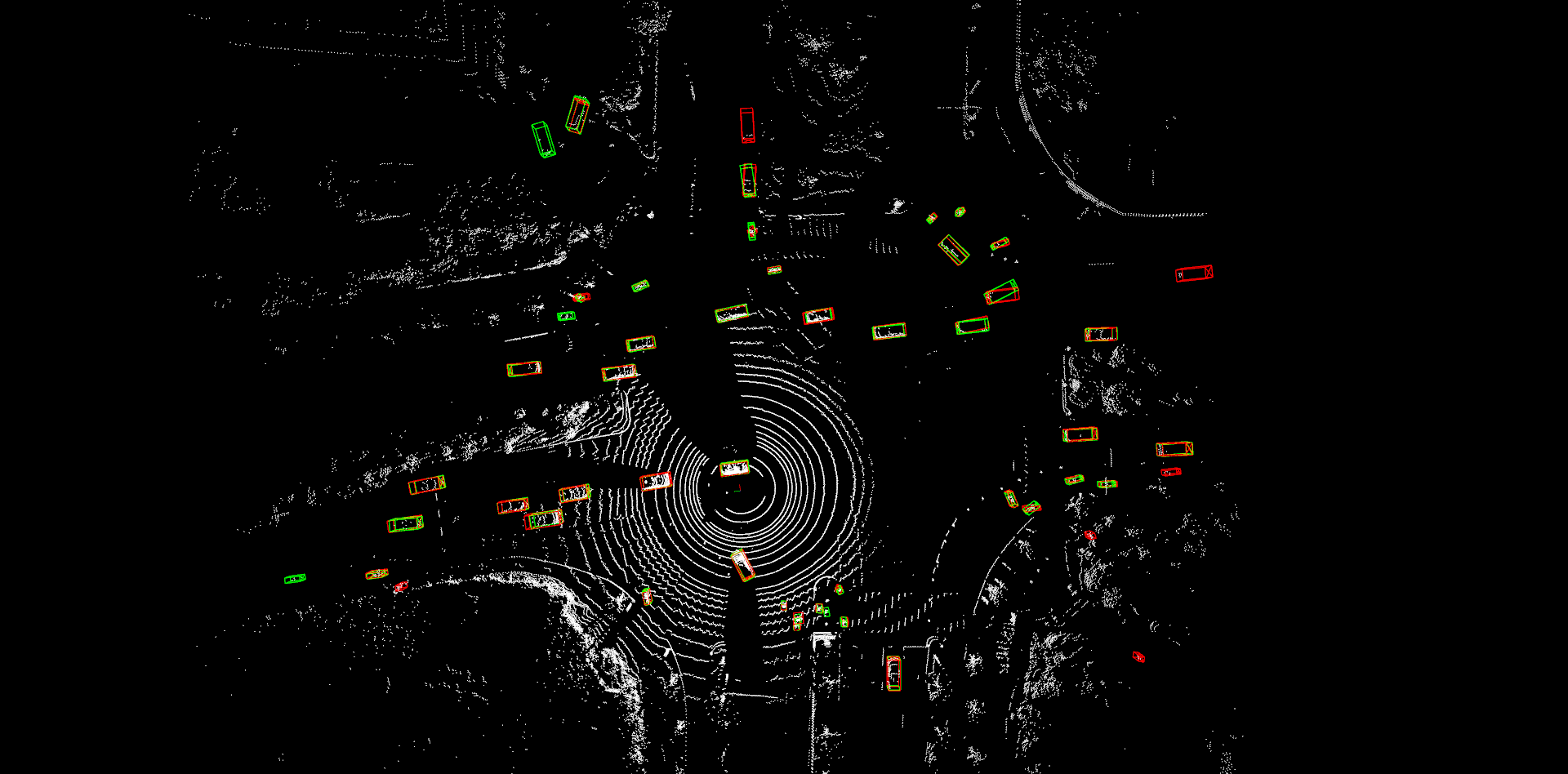} \\
    \includegraphics[width=0.95\linewidth]{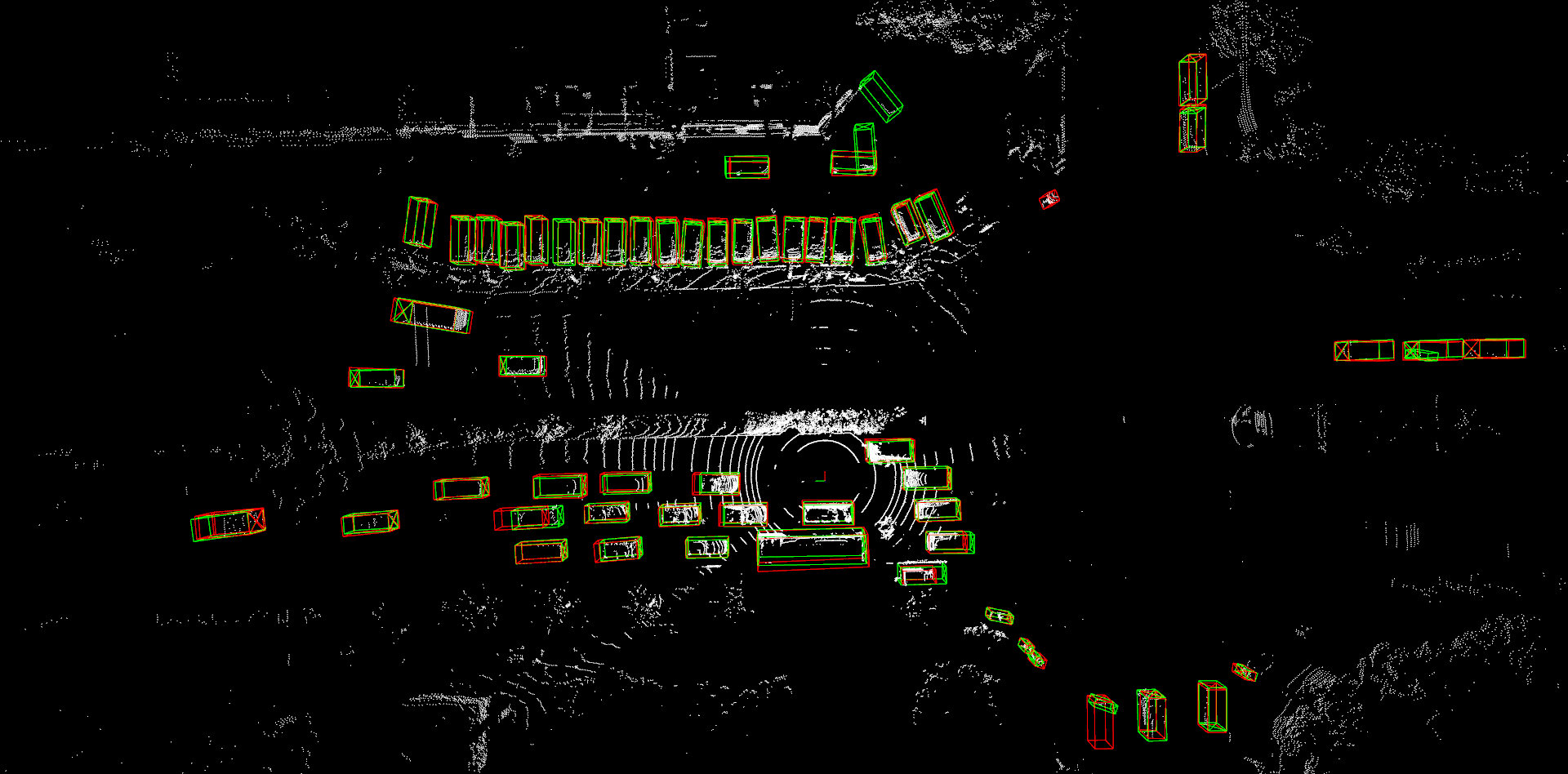}
    \caption{\textbf{Qualitative results on the ONCE  dataset~\cite{jMao_2021_ONCE}.} We depict ground truth and predictions as boxes colored in red and green for several exemplary scenes.} 
    \label{fig:supp_once_qua2}
\end{figure*}

\fi

\end{document}